\documentclass[preprint,12pt,authoryear]{elsarticle}
\usepackage{amssymb}
\usepackage[utf8]{inputenc}
\usepackage{geometry}
\usepackage{float}
\usepackage{array}
\usepackage{graphicx}
\usepackage{natbib}
\usepackage{tabularx}
\usepackage{booktabs}
\usepackage{graphicx}

\usepackage{amsmath}
\usepackage{hyperref}
\usepackage{algorithm}
\usepackage{algpseudocode}
\usepackage{booktabs}
\usepackage{subcaption}
\usepackage{multirow}
\usepackage{CJKutf8}
\usepackage{enumitem} 
\usepackage[figuresright]{rotating} 

\usepackage{placeins}

\hypersetup{
    colorlinks = true,
    citecolor = black,
    linkcolor = black,
    urlcolor = black 
}

{ \bgroup
	\addtolength\abovedisplayshortskip{#1}
	\addtolength\abovedisplayskip{#1}
	\addtolength\belowdisplayshortskip{#1}
	\addtolength\belowdisplayskip{#1}}
{\egroup\ignorespacesafterend}

\journal{*******}

\usepackage{adjustbox}

\begin{document}

\begin{frontmatter}
    \title{Language-Grounded Semantic Target Navigation for Autonomous Surface Vehicles\tnoteref{label1}}

    \author[inst1]{Yuqing Lin}
    \ead{yuqing003@e.ntu.edu.sg}

    
    \author[inst1]{Youngrong Kim\corref{cor1}}
    \ead{youngrong.kim@ntu.edu.sg}
    \cortext[cor1]{Corresponding author}
    
    \affiliation[inst1]{
        organization={School of Civil and Environmental Engineering, Nanyang Technological University},
        country={Singapore}
    }

    \begin{abstract}
    Autonomous Surface Vehicles (ASVs) are increasingly expected to operate in ports and harbour environments, where operators may specify navigation targets through language-based descriptions rather than predefined coordinates or fixed target identifiers. However, existing ASV navigation methods mainly execute predefined geometric goals or task-specific objectives and give limited attention to language-grounded target specification. This study proposes Semantically Grounded Navigation (SGNav), a framework that enables an ASV to identify and approach a maritime target from an operator-provided description. SGNav integrates text-guided semantic grounding, harbour-aware candidate filtering, CLIP-based semantic verification, grounded target control-state construction, and Proximal Policy Optimisation-based closed-loop control. It grounds the target description in onboard RGB observations, suppresses visually or semantically irrelevant distractors, and converts the selected target into a compact control-oriented representation for policy execution. Experiments in simulated port environments show that SGNav achieves success rates of $97.0\pm1.2\%$, $92.0\pm1.5\%$, and $90.0\pm1.8\%$ across three representative target-reaching tasks, with semantic target accuracy above $97\%$ and wrong-target rates below $3\%$. SGNav also maintains $97.7$--$98.7\%$ success rates across held-out port layouts. In the Task~3 ablation study, removing harbour-aware filtering or semantic consistency reduces the success rate to $40.4\pm2.6\%$ and $50.4\pm3.1\%$, respectively. These findings demonstrate the importance of semantic grounding, harbour-aware filtering, and semantic verification for reliable language-grounded ASV navigation. These results indicate that the proposed perception-to-control framework can support language-grounded target approach manoeuvres of ASV under the complex port environments.
    \end{abstract}
    
    \begin{keyword}
    Maritime Autonomous Systems \sep Semantic Grounding \sep Autonomous Surface Vehicles \sep Harbour navigation \sep Vision-Language Navigation
    \end{keyword}

\end{frontmatter}

\section{Introduction}

Maritime autonomous surface vehicles (ASVs) are increasingly deployed for port inspection, environmental monitoring, and harbour service operations. These missions require reliable manoeuvring and target identification in confined waters~\citep{qiao2023survey,lin2025machine}. Compared with conventional crewed vessels, ASVs can reduce human exposure to repetitive or hazardous maritime tasks. However, autonomous navigation in harbour environments remains challenging because an ASV must respond to vessel traffic, avoid obstacles, and distinguish navigational aids from vessels and fixed port structures. Prior studies have made substantial progress in maritime path planning and collision avoidance~\citep{singh2018constrained,liu2025hybrid}, while learning-based methods have further improved ASV decision-making and control under dynamic conditions~\citep{luo2025lstm,qu2025collaborative}. Nevertheless, most existing methods focus on executing a navigation objective that has already been provided in a machine-readable form, such as a coordinate, reference path, target category, or fixed object identifier.

This assumption creates a a mismatch between operator-provided target descriptions and the goal representations accepted by most ASV controllers. In Vessel Traffic Service (VTS), pilotage, and other operator-supervised maritime operations, navigational information and operational messages are commonly exchanged through voice communication using standardised maritime terminology and phraseology~\citep{imo2001smcp,iala2022g1132}. However, an autonomous navigation system do not generally convert such descriptions directly into control-ready target states. The intended task or target must first be translated into coordinates, waypoints, predefined routes, or fixed target identifiers that can be processed by the vehicle controller. This additional conversion limits operational flexibility and requires prior knowledge of the target location or an established mapping between the target and its machine-readable identifier.

The limitation is particularly relevant in local inspection and target-approach operations, where an operator may recognise a target from its semantic or visual characteristics without knowing its precise coordinates or predefined identifier. For example, a temporary inspection target may be described as "the yellow buoy near the dock", "the green navigation marker", or "the red floating object beside the vessel". Such descriptions identify the target through attributes including object category, colour, appearance, and spatial context. Under a conventional navigation workflow, the operator must first identify the object, determine its location, and manually convert it into a waypoint or target identifier before the ASV can begin the approach manoeuvre. This workflow requires prior target localization and manual goal encoding, which may not be available for temporary or previously unmapped targets.

Language-grounded target navigation provides a potential mechanism for reducing this gap. A language-based target description can be associated with the ASV's onboard visual observations to identify the intended maritime object. The resulting grounded target can subsequently be transformed into a navigation objective for closed-loop control. Although the present study does not address speech recognition or the interpretation of complete VTS manoeuvring instructions, it investigates an important intermediate capability for future maritime autonomy: enabling an ASV to identify and approach a visually observable target specified through semantic language. 

Recent advances in Vision-and-Language Navigation (VLN), semantic navigation, open-vocabulary visual grounding, and language-conditioned robotics provide useful foundations for addressing this problem. VLN studies have demonstrated that natural language can provide a flexible interface for specifying navigation goals and routes~\citep{anderson2018vision,gu2022vision}. More recent foundation-model-based navigation methods further demonstrate the potential of vision-language models, language models, and large-scale robot datasets for goal-conditioned navigation~\citep{shah2023lmnav,sridhar2024nomad}. Meanwhile, open-vocabulary detection and visual grounding methods provide practical mechanisms for associating textual concepts with visual objects or image regions~\citep{radford2021learning,liu2024grounding}. However, these methods are not directly sufficient for language-grounded semantic target navigation in maritime environments. Most language-conditioned navigation methods are developed for indoor environments, household robots, ground vehicles, or terrestrial outdoor scenarios. Generic visual grounding models, in contrast, are primarily designed to localise semantic concepts in images and are not explicitly optimised for closed-loop ASV navigation under harbour-specific distractors, including water reflections, wave regions, vessels, piers, non-target buoys, and visually similar maritime objects.

Language-grounded semantic target navigation for ASVs therefore requires more than directly applying a vision-language model to an onboard camera image. The system must first ground the natural-language target description to the correct maritime object, reject visually or semantically similar distractors, and then transform the grounded target into a control-oriented representation that can be processed by a navigation policy. This coupling between semantic grounding and closed-loop control is important because the output of a visual grounding model, such as a bounding box or semantic similarity score, does not directly constitute an executable navigation objective. A compact navigation-oriented target representation is therefore required to connect language-based target specification, onboard visual perception, and ASV control. Accordingly, this study addresses how ASVs can resolve an operator-specified maritime target from onboard observations and use the resolved target for closed-loop control.

This study proposes a language-grounded target navigation framework, referred to as Semantically Grounded Navigation (SGNav), for ASVs operating in port environments. SGNav uses text-guided semantic grounding to associate an operator-provided target description with candidate objects in onboard red-green-blue (RGB) camera observations. Harbour-specific candidate filtering is then applied to suppress irrelevant maritime regions, followed by vision-language semantic verification to identify the candidate that best matches the target description. The grounded target is subsequently converted into a compact grounded target control state, which is used by a Proximal Policy Optimisation (PPO)-based policy to generate closed-loop ASV control actions. In this way, SGNav establishes a perception-to-control pipeline from language-based target specification and maritime visual grounding to target-conditioned ASV control.

The main contributions of this paper are summarised as follows:
\begin{itemize}
\item A language-grounded semantic target navigation problem is formulated for maritime ASVs, where the navigation target is specified through an operator-provided semantic description rather than a predefined coordinate, reference trajectory, or fixed target identifier.

\item SGNav is developed to link language-based target descriptions, onboard RGB camera observations, maritime target grounding, and downstream ASV control through harbour-aware filtering, CLIP-based verification, and grounded target control-state construction.

\item Experiments across target-reaching tasks, instruction variants, distractor settings, held-out port layouts, and ablation settings are used to assess SGNav under different semantic, visual, and spatial conditions.
\end{itemize}

The remainder of this article is organised as follows. Section~\ref{sec:2} reviews related work on language-guided navigation, semantic grounding, and autonomous maritime control. Section~\ref{sec:3} presents the proposed SGNav framework. Section~\ref{sec:4} describes the experimental design and implementation settings. Section~\ref{sec:5} reports the experimental results and analysis. Finally, Section~\ref{sec:6} concludes this study and outlines future research directions.

\section{Related Work}
\label{sec:2}

\subsection{Vision-Language and Language-Conditioned Navigation}

Conventional robotic navigation typically represents goals using coordinates, waypoints, reference trajectories, or target images~\citep{paden2016survey,saravanakumar2011waypoint}. VLN extends this formulation by requiring an embodied agent to interpret natural-language instructions from visual observations and translate them into navigation actions~\citep{anderson2018vision,gu2022vision}. Early VLN studies mainly focused on indoor instruction following and improved cross-modal alignment, progress estimation, action prediction, and memory-based reasoning~\citep{fried2018speaker,ma2019self,wang2019reinforced,ku2020room}.

Recent studies increasingly incorporate Vision-Language Models (VLMs), Large Language Models (LLMs), and robotic foundation models into navigation. LM-Nav combines language parsing, visual-language matching, and a pretrained navigation policy to execute long-horizon natural-language instructions without requiring language-annotated robot trajectories~\citep{shah2023lmnav}. GNM, ViNT, and NoMaD learn generalisable navigation behaviours from heterogeneous robot trajectories and support cross-environment or cross-embodiment transfer~\citep{shah2023gnm,shah2023vint,sridhar2024nomad}. Other approaches investigate LLM-based navigation reasoning, subgoal generation, multimodal target specification, semantic memory, language-conditioned object navigation, and VLM-based outdoor navigation~\citep{zhou2024navgpt,zhou2024navgpt2,khanna2024goat,hirose2025lelan,wang2026expand,elnoor2025vlm}.

These studies demonstrate the potential of language and foundation models as flexible navigation interfaces. However, they are mainly designed for indoor scenes, household robots, ground vehicles, or terrestrial outdoor environments. Their assumptions do not directly address maritime ASVs, where targets may appear small or distant and must be distinguished under water reflections, partial occlusion, vessel clutter, and visually similar harbour objects. Moreover, many existing methods focus on route-level instruction following, visual-goal navigation, or discrete subgoal reasoning, whereas the task considered in this study requires an ASV to identify a semantically described maritime target from onboard RGB observations and continuously navigate towards it.

\subsection{Semantic Grounding and Control-Oriented Representation}

Semantic and object-goal navigation require an agent to reach a target defined by semantic identity rather than a known geometric position. Existing methods use semantic maps, object relations, scene graphs, knowledge graphs, and structured memory to support target search and long-horizon reasoning~\citep{chaplot2020object,jiang2023learning,du2020learning,kiran2022spatial,wang2024goal,luo2024learning,wang2021structured,wang2023gridmm,yang2024hogn}. LLM-based methods such as SayPlan further connect language-level planning with structured scene representations to generate executable robot plans~\citep{rana2023sayplan}. However, these approaches generally assume predefined object categories, prior semantic structures, or persistent spatial maps.

Visual grounding provides a direct mechanism for associating textual target descriptions with image regions. Open-vocabulary detectors such as MDETR, GLIP, OWL-ViT, and GroundingDINO support object localisation from category names or free-form expressions~\citep{kamath2021mdetr,li2022grounded,zhang2022glipv2,minderer2022simple,kuo2023fvlm,kim2023region,minderer2023scaling,liu2024grounding}. GroundingDINO is particularly suitable for text-conditioned candidate generation~\citep{zhang2022dino,liu2024grounding}, while CLIP provides aligned visual-textual representations for zero-shot recognition and semantic matching~\citep{radford2021learning,zhong2022regionclip,luddecke2022clipseg,rao2022denseclip}. Together, they support a two-stage process in which language-conditioned candidates are generated and subsequently verified against the complete semantic description.

However, grounding outputs such as bounding boxes, confidence scores, or semantic similarity values are not directly executable by an ASV controller. The selected target must be filtered against harbour-specific distractors and transformed into a compact representation of its relative image position, scale, and visibility. Related engineering informatics studies similarly demonstrate the importance of converting human-interpretable semantic information into machine-processable representations ~\citep{khairuddin2015review,lee2007constrained,shalal2015orchard}. Semantic SLAM, BIM, BIM-semantic maps, and digital twins integrate geometric and contextual information for localisation, planning, semantic interaction, and intelligent control~\citep{yang2024enhanced,chen2022pathfinding,xue2021semantic,kim2022bim,yang2024digital,berg2025digital}. Unlike these approaches, which generally depend on predefined maps or structured models, SGNav derives a control-ready target state directly from an operator-provided language description and onboard RGB observations.

\subsection{Autonomous Maritime Navigation and Reinforcement Learning Control}

Autonomous maritime navigation has traditionally focused on route planning, path following, obstacle avoidance, collision avoidance, and motion control. Conventional ASV navigation methods can be broadly grouped into waypoint- or graph-search-based planning, optimisation-based trajectory generation, Model Predictive Control, and rule-aware collision avoidance~\citep{yu2021usv,schoener2022anytime,wu2024efficient}. These approaches provide effective solutions for geometric navigation and obstacle avoidance when the navigation objective, environmental model, or reference route has already been specified~\citep{singh2018constrained,liu2025hybrid}. However, they usually assume that the target or goal is available in a machine-readable form, such as a waypoint, reference trajectory, or predefined object identifier. They therefore do not directly address how an ASV should identify a navigation target described through semantic language from onboard observations.

Deep Reinforcement Learning (DRL) has further improved adaptive maritime control for obstacle avoidance, collaborative navigation, multi-USV coordination, maritime risk assessment, and energy-efficient harbour craft operation~\citep{luo2025lstm,qu2025collaborative,zhang2024multi,maidana2023risk,lin2025multiple,lin2025multi}. Although these methods improve decision-making under dynamic obstacles and environmental disturbances, their objectives are generally encoded through coordinates, waypoints, desired headings, reference trajectories, relative target states, operational indicators, or task-specific reward functions.

Recent studies have begun to incorporate LLMs and VLMs into maritime navigation and mission execution. LLM4SAC uses LLM-generated guidance to improve reinforcement learning for a predefined USV docking task~\citep{xu2025llm4sac}. AI Captain employs a conversational LLM for high-level mission planning and replanning, while conventional guidance and control modules execute the resulting behaviours~\citep{christensen2025aicaptain}. USV-3.0 combines language-conditioned VLM perception, human-in-the-loop learning, and spatio-temporal memory to retrieve demonstrated maritime behaviours~\citep{salgado2026usv3}, whereas Semantic Lookout uses VLM-based scene understanding to identify semantic hazards and select safety manoeuvres from constrained candidates~\citep{christensen2026foundation}. These studies demonstrate the potential of foundation models for maritime autonomy, but they focus on docking guidance, mission decomposition, behaviour retrieval, or safety manoeuvre selection rather than open-vocabulary semantic target navigation.

More fundamentally, existing maritime DRL methods mainly address how to control an ASV after the target has already been specified. Their policies are trained using geometric states, predefined goals, and task-specific reward functions, and therefore cannot independently determine which observed object corresponds to an operator's semantic description. In a cluttered harbour, an incorrectly selected buoy or vessel may still produce a geometrically valid target state, leading to smooth and collision-free navigation towards the wrong object. Such failures may not be fully resolved by improving the downstream policy alone, because the ambiguity arises from target identity rather than only from control quality. This motivates the use of a semantic grounding mechanism to provide target-specific information before closed-loop control.

\subsection{Research Gap}

Existing research provides complementary foundations for language-conditioned navigation, visual grounding, engineering semantic representation, and autonomous maritime control, but these capabilities remain insufficiently integrated. General language-navigation methods are predominantly developed outside the maritime domain. Although open-vocabulary grounding methods can associate textual queries with image regions, they do not explicitly address harbour-specific distractors or provide control-oriented target states for ASV navigation. Structured engineering information systems support spatial modelling and operational decision-making, but they generally rely on predefined maps, digital models, or manually maintained object information. Maritime DRL methods mainly focus on policy learning and collision avoidance after the intended target has already been specified.

This leaves a gap between language-based maritime target specification and executable ASV control. In practical target-approach scenarios, an operator may describe "the green buoy", "the yellow patrol boat", or "the safe-water mark", while the ASV must identify the intended object among water reflections, partial occlusion, vessel clutter, non-target markers, and visually similar distractors. SGNav addresses this gap by grounding the operator-provided target description in onboard observations, verifying the target under harbour-specific interference, and converting it into a grounded target control state for closed-loop navigation.

\section{Methodology}
\label{sec:3}

\subsection{Overview}

Building on the limitations identified in Section~\ref{sec:2}, this study proposes SGNav, a language-grounded semantic target navigation framework for ASVs operating in visually cluttered harbour environments. SGNav follows a semantic-to-control decomposition: it first identifies the maritime object that best matches an operator-provided target description and then converts the grounded target into a compact state for continuous ASV control. This design separates semantic target identification from low-level motion execution, which are typically treated as independent assumptions in conventional navigation systems.

At each time step $t$, the ASV receives an operator-provided target description $g$, an onboard visual observation $I_t$, interpreted as an RGB frame, and an environment-related navigation observation $o_t^{\mathrm{env}}$. SGNav grounds $g$ in $I_t$ to obtain a verified target region $b_t^{*}$, encodes this region as a grounded target control state $z_t$, and combines $z_t$ with $o_t^{\mathrm{env}}$ to generate the continuous ASV control action $u_t$:
\[
(g,I_t)
\rightarrow
b_t^{*}
\rightarrow
z_t
\rightarrow
u_t.
\]

The novelty of SGNav lies in the coordinated integration of open-vocabulary target grounding, harbour-aware distractor suppression, semantic verification, and control-oriented target representation within a unified closed-loop navigation framework. Rather than directly passing generic grounding outputs to a control policy, SGNav resolves the intended maritime target and constructs a compact interface between semantic perception and downstream action execution.

\begin{figure}
\centering
\includegraphics[width=\linewidth]{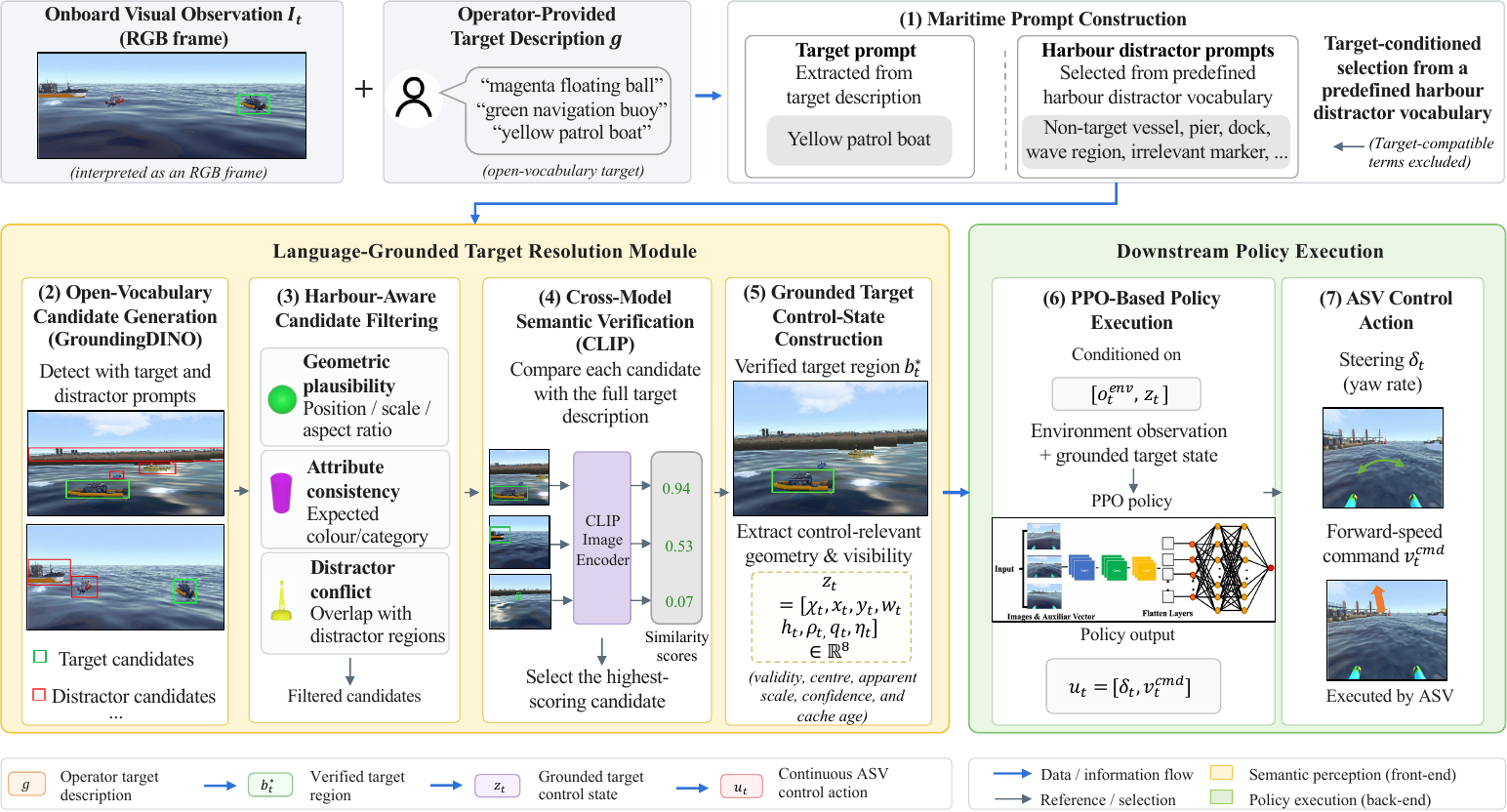}
\caption{Overall framework of SGNav for language-grounded semantic target navigation. The framework consists of a language-grounded target resolution front-end and a downstream policy execution back-end.}
\label{fig:method}
\end{figure}

Figure~\ref{fig:method} summarises the data flow from the operator-provided target description and onboard observations to target resolution and policy execution. Section~\ref{subsec:semantic_frontend} describes the language-grounded target resolution module, and Section~\ref{subsec:policy_execution} presents the downstream policy execution module.

\subsection{Language-Grounded Target Resolution}
\label{subsec:semantic_frontend}

The language-grounded target resolution module is the core perception component of SGNav. It resolves an operator-provided target description against onboard visual observations and converts the verified target into a compact grounded target control state for downstream navigation. As shown in Fig.~\ref{fig:method}, this module consists of maritime prompt construction, GroundingDINO-based open-vocabulary candidate generation, harbour-aware candidate filtering, CLIP-based semantic verification, and grounded target control-state construction.

\subsubsection{Maritime Prompt Construction}

Given an operator-provided target description $g$, SGNav constructs a target prompt set $\mathcal{P}^{+}(g)$ and a harbour distractor prompt set $\mathcal{P}^{-}(g)$. The prompt construction process is rule-based rather than generated by an LLM. Specifically, the target description is parsed into target-related attributes, such as object category, colour, and optional contextual terms, and is then expanded using predefined target-expression templates. The target prompt set is defined as:
\begin{equation}
\mathcal{P}^{+}(g)
=
\{p^{+}_{1},p^{+}_{2},\ldots,p^{+}_{M}\},
\end{equation}
where each $p^{+}_{j}$ denotes the intended target using either the original wording or a predefined semantically equivalent expression. For example, the description "yellow patrol boat" may produce prompts such as "yellow patrol boat" and "yellow boat", while "green navigation buoy" may produce prompts such as "green navigation buoy" and "green buoy".

To reduce false grounding in cluttered harbour scenes, SGNav also uses a predefined harbour distractor vocabulary $\mathcal{V}_{\mathrm{harbour}}$, which contains common non-target concepts such as non-target vessels, piers, docks, wave regions, land structures, irrelevant markers, and incompatible buoy classes. The initial distractor prompt set is selected from this vocabulary:
\begin{equation}
\mathcal{P}^{-}_{0}
=
\{p^{-}_{1},p^{-}_{2},\ldots,p^{-}_{N}\},
\quad
p^{-}_{k}\in\mathcal{V}_{\mathrm{harbour}}.
\end{equation}

Before the distractor prompts are used, SGNav removes target-compatible terms from the negative set. This step prevents the intended target or its semantic parent category from being treated as a negative class. For example, when the target is "green navigation buoy", generic or compatible terms such as "buoy" and "navigation marker" are excluded from the negative prompt set, while incompatible concepts such as "ship", "pier", "dock", and "wave region" are retained. Formally, the final harbour distractor prompt set is defined as:
\begin{equation}
\mathcal{P}^{-}(g)
=
\left\{
p^{-}\in\mathcal{P}^{-}_{0}
\mid
\operatorname{sim}_{\mathrm{text}}(p^{-},g)<\tau_{\mathrm{excl}}
\right\},
\end{equation}
where $\operatorname{sim}_{\mathrm{text}}(\cdot)$ denotes the text-level semantic similarity between a distractor term and the current target description, and $\tau_{\mathrm{excl}}$ is the exclusion threshold for removing target-compatible distractor terms. In implementation, $\operatorname{sim}_{\mathrm{text}}(\cdot)$ is computed using the cosine similarity between CLIP text embeddings, and manually defined synonym and parent-category rules are also used to avoid obvious semantic conflicts. If all distractor terms are excluded for a specific target, SGNav performs candidate generation using only the positive target prompts and skips the distractor-conflict filtering term for that frame. This target-conditioned selection mechanism clarifies that the harbour distractor prompts are predefined domain distractors filtered according to the current target description, rather than arbitrary or automatically hallucinated negative prompts.

Table~\ref{tab:prompt_construction} provides examples of the rule-based prompt construction used in the experiments. The table is not intended to exhaust all possible maritime expressions, but to clarify how target prompts and harbour distractor prompts are formed in the evaluated tasks.

\begin{table}[t]
\centering
\caption{Examples of rule-based prompt construction in SGNav. Target prompts are generated from the operator-provided description and predefined target-expression templates, while harbour distractor prompts are selected from a predefined vocabulary after removing target-compatible terms.}
\label{tab:prompt_construction}
\begin{tabular}{p{0.22\linewidth} p{0.33\linewidth} p{0.35\linewidth}}
\toprule
Target description $g$ & Target prompts $\mathcal{P}^{+}(g)$ & Harbour distractor prompts $\mathcal{P}^{-}(g)$ \\
\midrule
Magenta Floating Marker &
magenta floating marker, magenta floating ball, floating magenta object &
vessel, ship, pier, dock, wave region, land structure, irrelevant buoy \\
\midrule
green navigation buoy &
green navigation buoy, green buoy, green navigation marker &
vessel, ship, pier, dock, wave region, land structure, irrelevant marker, incompatible buoy class \\
\midrule
yellow patrol boat &
yellow patrol boat, yellow boat, patrol vessel &
buoy, pier, dock, wave region, land structure, irrelevant marker, non-target vessel \\
\bottomrule
\end{tabular}
\end{table}

\subsubsection{Open-Vocabulary Candidate Generation}

Given the onboard visual observation $I_t$, SGNav uses GroundingDINO to generate text-conditioned visual hypotheses. GroundingDINO supports open-vocabulary object detection using category names or referring expressions as textual inputs~\citep{liu2024grounding,yao2026improving}, making it suitable for language-conditioned maritime target grounding.

The target prompt set produces candidate target boxes:
\begin{equation}
\mathcal{B}^{+}_{t}
=
\mathrm{GDINO}(I_t,\mathcal{P}^{+}(g)),
\end{equation}
where each candidate $b_i^{+}\in\mathcal{B}^{+}_{t}$ contains a bounding box and a detection confidence score. In parallel, the harbour distractor prompt set produces distractor hypotheses:
\begin{equation}
\mathcal{B}^{-}_{t}
=
\mathrm{GDINO}(I_t,\mathcal{P}^{-}(g)).
\end{equation}
The two candidate sets provide the raw target and target-conditioned distractor hypotheses for subsequent harbour-aware filtering and semantic verification.

\subsubsection{Harbour-Aware Candidate Filtering}

The raw GroundingDINO detections are refined using harbour-aware candidate filtering. This stage applies lightweight domain priors to suppress candidates that are inconsistent with the expected target appearance or strongly associated with known harbour distractors. The filtering follows common post-processing practices in object detection and open-vocabulary region recognition, while adapting them to the harbour navigation context~\citep{wu2023cora,son2024teacher}.

For each target candidate $b_i^{+}$, SGNav first applies a geometric plausibility filter:
\begin{equation}
g_i =
\mathbb{I}
\left[
b_i^{+}\in\Omega_{\mathrm{geo}}
\right],
\end{equation}
where $\Omega_{\mathrm{geo}}$ denotes the valid geometric range defined by bounding-box position, scale, and aspect ratio. This filter removes candidates that are unlikely to correspond to valid maritime targets, such as extremely large background regions or implausible water-surface detections.

SGNav then applies an attribute-consistency filter:
\begin{equation}
h_i =
\mathbb{I}
\left[
\rho_{\mathrm{attr}}(b_i^{+},g)>\tau_{\mathrm{attr}}
\right],
\end{equation}
where $\rho_{\mathrm{attr}}(b_i^{+},g)$ measures the consistency between the candidate region and target attributes extracted from $g$, such as expected colour or category, and $\tau_{\mathrm{attr}}$ is the corresponding threshold. This term is written as an attribute-consistency filter rather than only a colour filter because some target descriptions may specify object category or visual type rather than colour alone.

To account for harbour distractor candidates, SGNav further computes the distractor conflict score for each target candidate:
\begin{equation}
d_i =
\max_{b_j^{-}\in\mathcal{B}^{-}_{t}}
\mathrm{IoU}(b_i^{+},b_j^{-}),
\end{equation}
where $b_i^{+}$ denotes a target candidate and $b_j^{-}$ denotes a distractor candidate generated from the harbour distractor prompt set. A high conflict score indicates that a target candidate strongly overlaps with a region associated with a harbour distractor. If no distractor candidate is generated, $d_i$ is set to zero.

The filtered target candidate set is then obtained as:
\begin{equation}
\hat{\mathcal{B}}_{t}
=
\left\{
b_i^{+}\in\mathcal{B}^{+}_{t}
\mid
g_i=1,\ h_i=1,\ d_i<\tau_{\mathrm{conf}}
\right\}.
\end{equation}
Here, $\tau_{\mathrm{conf}}$ is the distractor-conflict threshold. A target candidate is retained only when it satisfies geometric plausibility, attribute consistency, and does not strongly overlap with a distractor candidate.

\subsubsection{CLIP-Based Semantic Verification}

After harbour-aware filtering, SGNav uses CLIP as a second-stage semantic verifier. CLIP learns aligned image and text representations from natural-language supervision, allowing cropped candidate regions to be compared with textual descriptions~\citep{radford2021learning,yu2023fusing}. This verification step is used to select the candidate that best matches the target prompt set derived from the operator-provided description, rather than simply accepting the highest-confidence GroundingDINO detection.

For each remaining candidate $b_i\in\hat{\mathcal{B}}_{t}$, SGNav crops the candidate image region $c_i=I_t[b_i]$ and encodes it using the CLIP image encoder $f_{\mathrm{img}}(\cdot)$. The target similarity score is computed as:
\begin{equation}
s_i
=
\max_{p^{+}\in\mathcal{P}^{+}(g)}
\cos
\left(
f_{\mathrm{img}}(c_i),
f_{\mathrm{text}}(p^{+})
\right),
\end{equation}
where $f_{\mathrm{text}}(\cdot)$ denotes the CLIP text encoder. The maximum operation allows each candidate to be matched against multiple target expressions generated from the original description $g$, improving robustness to wording variation while preserving the intended target semantics. The verified target region is selected as:
\begin{equation}
b_t^{*}
=
\arg\max_{b_i\in\hat{\mathcal{B}}_{t}} s_i,
\quad
\mathrm{s.t.}\quad
s_i>\tau_s .
\end{equation}
Here, $\tau_s$ is the semantic verification threshold. If no candidate satisfies this threshold, the target is treated as temporarily unobserved. This design keeps the CLIP verification consistent with Fig.~\ref{fig:method}, where candidate crops are compared with the complete target description and the highest-scoring valid candidate is selected.

\subsubsection{Grounded Target Control-State Construction}

The verified target region $b_t^{*}$ is converted into an 8-dimensional grounded target control state:
\begin{equation}
z_t =
[\chi_t,x_t,y_t,w_t,h_t,\rho_t,q_t,\eta_t]
\in\mathbb{R}^{8}.
\end{equation}
Here, $\chi_t$ is a validity indicator denoting whether a target is currently observed, $x_t$ and $y_t$ denote the normalised target centre, $w_t$ and $h_t$ denote the normalised bounding-box width and height, $\rho_t$ denotes the normalised bounding-box area, $q_t$ denotes the grounding confidence, and $\eta_t$ denotes the normalised cache age of the latest valid target observation.

When a valid target is observed, the normalised target centre and apparent-scale terms are computed as:
\begin{equation}
x_t=\frac{x_{\mathrm{centre}}(b_t^{*})}{W},
\quad
y_t=\frac{y_{\mathrm{centre}}(b_t^{*})}{H},
\end{equation}
\begin{equation}
w_t=\frac{w(b_t^{*})}{W},
\quad
h_t=\frac{h(b_t^{*})}{H},
\quad
\rho_t=\frac{w(b_t^{*})h(b_t^{*})}{WH},
\end{equation}
where $W$ and $H$ are the image width and height. The bounding-box centre provides an image-plane alignment cue, while the width, height, and area provide apparent-scale cues. SGNav does not infer absolute metric distance from $y_t$ or from a single monocular image observation, because objects with different physical sizes may produce similar image-plane positions or scales. Therefore, $z_t$ should be interpreted as a compact control-oriented visual target representation rather than a metric 3D target localisation result. The bounding-box centre is not treated as a physical destination point; safety-aware approach behaviour is handled by the downstream policy through its reward design and termination conditions.

The grounding confidence combines the CLIP semantic similarity and the GroundingDINO detection confidence:
\begin{equation}
q_t
=
\sigma(\alpha s_t+\beta r_t^{\mathrm{det}}),
\end{equation}
where $s_t$ is the CLIP similarity score of the selected candidate, $r_t^{\mathrm{det}}$ is the GroundingDINO detection confidence, $\alpha$ and $\beta$ are weighting coefficients, and $\sigma(\cdot)$ denotes sigmoid normalisation. This confidence term provides a compact reliability cue for the downstream policy, which is consistent with grounding-based policy representations~\citep{jiang2024visual}.

To improve temporal stability, SGNav maintains a short-term target cache. When the target is temporarily lost due to occlusion, viewpoint change, or detection failure, the most recent valid grounded target control state can be reused for a limited number of steps. The cache age $\eta_t$ allows the policy to distinguish fresh target evidence from older cached information. If no valid target is observed and no valid cache is available, SGNav sets $\chi_t=0$ and uses a default null target state.

\subsection{Downstream Policy Execution}
\label{subsec:policy_execution}

The downstream policy execution module uses the grounded target control state produced by the language-grounded target resolution module to control the ASV. This module is separated from the perception front-end to make the role of SGNav clear: semantic perception resolves which visual object matches the operator-provided target description, while the policy learns how to translate the resulting target state and environment observation into continuous control commands.

\subsubsection{Policy Input Formulation}

The grounded target control state is concatenated with the environment-related navigation observation:
\begin{equation}
o_t^{\mathrm{policy}}
=
[o_t^{\mathrm{env}},z_t],
\end{equation}
where $o_t^{\mathrm{env}}$ denotes the original navigation-related observation used by the ASV controller, such as local motion states or environment feedback. This concatenation allows the policy to condition its behaviour on both low-level navigation information and the grounded target control state.

\subsubsection{ASV Action Execution}

The downstream policy generates the continuous ASV control action as:
\begin{equation}
u_t
=
\pi_{\theta}(o_t^{\mathrm{policy}})
=
[\delta_t,v_t^{\mathrm{cmd}}],
\end{equation}
where $\delta_t$ denotes the steering or yaw-rate command and $v_t^{\mathrm{cmd}}$ denotes the forward-speed command. PPO is adopted as the policy optimisation algorithm because of its stable clipped policy-gradient update~\citep{schulman2017proximal}. The PPO objective is written as:
\begin{equation}
L^{\mathrm{PPO}}(\theta)
=
\mathbb{E}_{t}
\left[
\min
\left(
r_t(\theta)\hat{A}_t,
\mathrm{clip}(r_t(\theta),1-\epsilon,1+\epsilon)\hat{A}_t
\right)
\right],
\end{equation}
where $r_t(\theta)$ is the probability ratio between the updated and old policies, $\hat{A}_t$ is the estimated advantage, and $\epsilon$ is the clipping parameter.

For safety-aware target approach, the ASV is trained to approach and stop within a prescribed stand-off region rather than collide with or physically contact the target. This is handled through both the reward design and the episode termination condition: target progress and image-plane alignment are rewarded, while collision, unsafe proximity, and excessive approach are penalised. Successful navigation is therefore defined by reaching the target-associated stand-off region while avoiding collision, not by reaching the centre of the detected bounding box. The stand-off and collision-related feedback are provided by the simulation environment during training and evaluation, rather than inferred solely from the image-plane coordinate $y_t$. In the current simulation, the stand-off criterion is specified according to the task setting. For larger real-world vessels or port structures, this safety region should be adjusted according to target class, physical size, and operational constraints.

It should be emphasised that PPO is not introduced as a new policy-learning contribution in SGNav. Instead, it serves as the downstream execution module that learns to use the grounded target control state produced by the proposed target resolution module. To summarise the complete inference process, Algorithm~\ref{alg:sgnav_short} presents the SGNav pipeline from operator-provided target description to ASV control output.

\begin{algorithm}[t]
\caption{SGNav Inference Procedure}
\label{alg:sgnav_short}
\begin{algorithmic}[1]
\Require Target description $g$, onboard visual observation $I_t$, environment observation $o_t^{\mathrm{env}}$
\Ensure ASV control action $u_t$

\State Construct target prompts $\mathcal{P}^{+}(g)$ from $g$
\State Select harbour distractor prompts $\mathcal{P}^{-}(g)$ from the predefined harbour distractor vocabulary, with target-compatible terms excluded
\State Generate the target candidate set $\mathcal{B}^{+}_{t}\leftarrow \mathrm{GDINO}(I_t,\mathcal{P}^{+}(g))$
\State Generate the distractor candidate set $\mathcal{B}^{-}_{t}\leftarrow \mathrm{GDINO}(I_t,\mathcal{P}^{-}(g))$
\State Filter target candidates using geometric plausibility, attribute consistency, and distractor-conflict constraints
\State Verify remaining candidates using CLIP semantic similarity
\If{a valid candidate is found}
    \State Select $b_t^{*}$ and construct $z_t=[\chi_t,x_t,y_t,w_t,h_t,\rho_t,q_t,\eta_t]$
\Else
    \State Use the short-term target cache or set $\chi_t=0$
\EndIf
\State Form policy input $o_t^{\mathrm{policy}}=[o_t^{\mathrm{env}},z_t]$
\State Compute ASV control action $u_t=\pi_{\theta}(o_t^{\mathrm{policy}})=[\delta_t,v_t^{\mathrm{cmd}}]$
\State \Return $u_t$
\end{algorithmic}
\end{algorithm}
\FloatBarrier

\section{Experiments}
\label{sec:4}

\subsection{Simulation Environment and Task Setting}
\label{subsec:simulation_environment}

\begin{figure}[H]
\centering
\includegraphics[width=\textwidth]{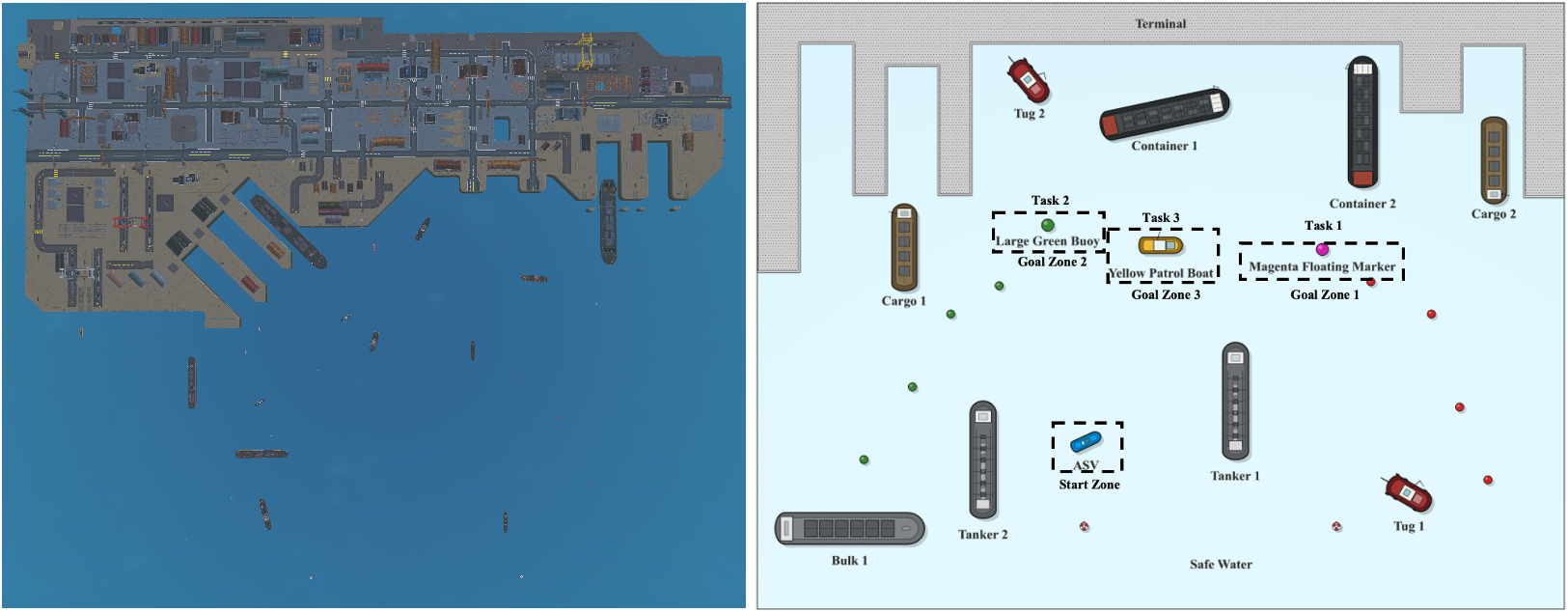}
\caption{
Simulation environment and target-reaching task configurations used in the benchmark. The left panel shows a Unity top-down overview of the simulated port environment, while the right panel provides a simplified schematic of the evaluated task layout. The schematic is not a one-to-one geometric reproduction of the Unity scene, but highlights the ASV start zone, representative distractors, navigation aids, and the three target-reaching tasks: Task~1 (magenta floating marker), Task~2 (large green buoy), and Task~3 (yellow patrol boat).
}
\label{fig:exps_123}
\end{figure}

The experiments were conducted in a high-fidelity Unity-based port simulation environment, as shown in Fig.~\ref{fig:exps_123}. The environment contains representative harbour elements, including terminals, quay facilities, piers, berth areas, navigation channels, open-sea traffic areas, navigation aids, and multiple vessel types. Together, these static and dynamic elements create visually cluttered maritime scenes. They introduce background interference, partial occlusion, water-surface variation, and nearby vessel traffic, which make semantic target grounding and closed-loop ASV navigation more challenging.

The benchmark includes three representative target-reaching tasks: Task~1, magenta floating marker navigation; Task~2, large green buoy navigation; and Task~3, yellow patrol boat navigation. All tasks share the same ASV start zone but use different instructed targets. As shown in Fig.~\ref{fig:exps_123}, Task~1 requires the ASV to approach a magenta floating marker near red buoy-like distractors, Task~2 requires navigation towards a large green buoy among smaller green buoys, and Task~3 requires approaching a yellow patrol boat surrounded by other vessel types. Task~1 serves as a controlled diagnostic case for colour- and shape-based grounding rather than a standard maritime navigation aid, while Tasks~2 and~3 evaluate more port-relevant size-aware buoy selection and vessel-class discrimination. Together, these tasks assess visual grounding, distractor rejection, and closed-loop navigation towards maritime targets.

Table~\ref{tab:smart_port_environment} summarises the main benchmark factors introduced by the port simulation environment. Rather than serving as an exhaustive asset list, the table highlights the semantic targets, distractors, port structures, traffic elements, and visual appearance variations that make language-grounded target navigation challenging.

\begin{table*}[t]
\centering
\caption{Key elements of the simulated port environment and their evaluation roles.}
\label{tab:smart_port_environment}
\renewcommand{\arraystretch}{1.12}
\setlength{\tabcolsep}{4pt}
\small
\begin{tabularx}{\textwidth}{p{0.22\textwidth} p{0.36\textwidth} X}
\toprule
Simulation factor & Representative elements & Evaluation role \\
\midrule

Semantic targets and distractors &
Navigation buoys, floating markers, patrol boats, safe-water marks &
Test target grounding and distractor rejection. \\

\addlinespace[0.3em]
Port structures &
Terminal, dock, pier, berth area, quay crane, turning basin, inbound channel &
Introduce occlusion, clutter, and spatial constraints. \\

\addlinespace[0.3em]
Vessels and traffic &
Container ships, cargo vessels, tugs, tankers, docked and moving vessels &
Increase target ambiguity and collision risk. \\

\addlinespace[0.3em]
Visual appearance variation &
Wave motion, water reflection, sunlight reflection, shadows &
Affect visibility, colour consistency, and grounding reliability. \\

\addlinespace[0.3em]
Scene ambiguity &
Background clutter, object similarity, partial occlusion &
Test semantic discrimination and target verification. \\

\bottomrule
\end{tabularx}
\end{table*}

\subsection{Experimental Protocol and Implementation Details}
\label{subsec:experimental_protocol}

Unless otherwise stated, a single shared PPO policy is trained across Tasks~1--3 for each method. In each episode, one ASV executes one uniformly sampled target-reaching task, and episodes from all tasks jointly optimise the same policy parameters $\pi_{\theta}$. Each method is trained for 9,000 episodes in total, corresponding to approximately 3,000 episodes per task, using three independent random seeds.

During evaluation, the learned policy from each seed is fixed and tested over 100 independent episodes per task. Results are reported as mean $\pm$ standard deviation over the three seeds. No task-specific fine-tuning is performed. An episode terminates when the ASV reaches the prescribed stand-off region, collides with an obstacle or target, reaches a wrong target, or exceeds the maximum time limit.

All experiments were conducted on a workstation equipped with a 13th Gen Intel Core i9-13980HX CPU, a mobile NVIDIA GeForce RTX 4080 GPU, and 62 GiB of RAM. GroundingDINO is used for open-vocabulary candidate generation, CLIP for semantic verification, and PPO for continuous ASV control.

Runtime was measured over 2,000 evaluation frames, including 1,000 frames from Task~1 and 1,000 frames from Task~3. As shown in Table~\ref{tab:runtime}, one full semantic refresh requires $335.5\pm18.5$ ms, corresponding to approximately 3.0 Hz. Since semantic perception is refreshed every five control steps and the latest grounded target state is cached between updates, the amortised cost is 67.5 ms per control step, corresponding to approximately 14.8 Hz. This amortised value represents average computational cost rather than worst-case synchronous latency.

The RGB observations, instruction inputs, ASV trajectories, and evaluation logs were generated in the Unity-based port simulator and no external image dataset was used. A minimal reproducibility package, including the SGNav execution code, task configurations, example evaluation materials, and a pre-built Unity executable, is provided at \url{https://github.com/linyqyq/sgnav-minimal-release}. 

\begin{table}[t]
\centering
\caption{Runtime of SGNav components measured over 2,000 evaluation frames.}
\label{tab:runtime}
\renewcommand{\arraystretch}{1.1}
\setlength{\tabcolsep}{3pt}
\small
\begin{tabular}{p{0.43\linewidth}cc}
\toprule
Component 
& \shortstack{Per refresh\\(ms)}
& \shortstack{Amortised\\(ms/step)} \\
\midrule

GroundingDINO 
& $310.0 \pm 18.0$
& $62.0$ \\

CLIP verification 
& $22.0 \pm 4.0$
& $4.4$ \\

Filtering and state construction 
& $3.0 \pm 1.0$
& $0.6$ \\

PPO inference 
& $0.5 \pm 0.1$
& $0.5 \pm 0.1$ \\

\midrule
Total 
& $335.5 \pm 18.5$
& $67.5$ \\

\bottomrule
\end{tabular}

\vspace{0.3em}
\begin{minipage}{0.95\linewidth}
\footnotesize
\textit{Note:} Semantic perception is updated every five control steps. The amortised cost divides the semantic perception cost by five, while PPO inference is counted at every step.
\end{minipage}
\end{table}

\subsection{Evaluation Metrics}
\label{subsec:metrics}

The experiments use a shared set of navigation and semantic grounding metrics. Let $N$ denote the number of evaluation episodes for a given method and setting. For episode $i$, let $S_i\in\{0,1\}$ indicate whether the ASV reaches the prescribed stand-off region of the instructed target without collision and within the maximum time limit. The Success Rate is defined as:
\begin{equation}
\mathrm{SR}
=
\frac{1}{N}
\sum_{i=1}^{N} S_i .
\end{equation}

Semantic Target Accuracy (STA) measures the frame-level accuracy of the grounding module at semantic refresh steps. Let $G_k\in\{0,1\}$ indicate whether the grounded target region at semantic refresh step $k$ corresponds to the instructed target. Then,
\begin{equation}
\mathrm{STA}
=
\frac{1}{K}
\sum_{k=1}^{K} G_k ,
\end{equation}
where $K$ is the total number of semantic refresh steps evaluated. 

Grounding Accuracy (GA) is used when evaluating the target grounding module. Let $G_i\in\{0,1\}$ indicate whether the grounded target region $b_t^{*}$ corresponds to the instructed target in episode $i$. Then,
\begin{equation}
\mathrm{GA}
=
\frac{1}{N}
\sum_{i=1}^{N} G_i .
\end{equation}

Wrong-Target Rate (WTR) measures the frequency of selecting or reaching a non-instructed target at the episode level. Let $W_i\in\{0,1\}$ indicate whether episode $i$ results in a wrong-target selection or wrong-target arrival:
\begin{equation}
\mathrm{WTR}
=
\frac{1}{N}
\sum_{i=1}^{N} W_i .
\end{equation}

For distractor-rich settings, Correct Target Rate (CTR) and Distractor Rejection Rate (DRR) are additionally reported. Let $C_i^{\mathrm{tar}}\in\{0,1\}$ indicate whether the instructed target is correctly selected or reached in episode $i$. The Correct Target Rate is:
\begin{equation}
\mathrm{CTR}
=
\frac{1}{N}
\sum_{i=1}^{N} C_i^{\mathrm{tar}} .
\end{equation}

Let $J_i$ denote the number of predefined distractors in episode $i$, and let $R_{ij}\in\{0,1\}$ indicate whether the $j$-th distractor in episode $i$ is rejected, namely it is not selected as the final grounded target and is not reached by the ASV. DRR is computed as:
\begin{equation}
\mathrm{DRR}
=
\frac{1}{\sum_{i=1}^{N}J_i}
\sum_{i=1}^{N}
\sum_{j=1}^{J_i}
R_{ij}.
\end{equation}

Collision Rate (CR) is reported as ColR to avoid confusion with COLREGs compliance. Let $K_i\in\{0,1\}$ indicate whether a collision occurs in episode $i$. Then,
\begin{equation}
\mathrm{ColR}
=
\frac{1}{N}
\sum_{i=1}^{N} K_i .
\end{equation}
ColR only measures collision frequency in the simulation environment. It does not evaluate COLREGs compliance, as rule-based maritime encounter compliance is outside the scope of this semantic target navigation study.

Completion Time (CT) measures the average simulated time, in seconds, required to complete successful episodes:
\begin{equation}
\mathrm{CT}
=
\frac{1}{|\mathcal{S}|}
\sum_{i\in\mathcal{S}} T_i ,
\end{equation}
where $\mathcal{S}=\{i\mid S_i=1\}$ is the set of successful episodes and $T_i$ denotes the simulated completion time of episode $i$ in seconds. If an episode is completed after $n_i$ environment steps with a simulation time interval $\Delta t$, then $T_i=n_i\Delta t$.

Path Efficiency (PE) evaluates the compactness of successful trajectories:
\begin{equation}
\mathrm{PE}
=
\frac{1}{|\mathcal{S}|}
\sum_{i\in\mathcal{S}}
\frac{L_i^{\mathrm{ref}}}{L_i},
\end{equation}
where $L_i$ is the travelled path length and $L_i^{\mathrm{ref}}$ is the reference shortest feasible path length from the start position to the target-associated stand-off region. A larger PE value indicates a more efficient trajectory.

\subsection{Experiment 1: Language-Grounded Semantic Target Navigation}
\label{subsec:exp1}

Experiment~1 evaluates the basic semantic target navigation capability of SGNav under the benchmark task settings defined in Section~\ref{subsec:simulation_environment}. The objective is to examine whether the ASV can ground an operator-provided target description, select the intended visual target from a cluttered harbour scene, and navigate to the prescribed stand-off region using a shared downstream policy.

The evaluation is conducted on the three target-reaching tasks shown in Fig.~\ref{fig:exps_123}: magenta floating marker navigation, large green buoy navigation, and yellow patrol boat navigation. These tasks cover a controlled colour-shape grounding case, size-aware buoy selection, and vessel-class discrimination, thereby testing whether SGNav can translate different language-grounded targets into effective closed-loop navigation.

SGNav is compared with three task-level baselines: No-Target PPO, Vision-Only PPO, and Oracle PPO. No-Target PPO receives no target-related information and represents unguided target-reaching behaviour. Vision-Only PPO uses visual and environment observations but does not include explicit language-grounded target resolution. Oracle PPO receives privileged ground-truth target information and serves as an upper-performance bound. In contrast, SGNav obtains the target state through language-grounded target resolution and uses the resulting grounded target control state for policy execution.

All methods follow the shared training and evaluation protocol described in Section~\ref{subsec:experimental_protocol}. The evaluation reports SR, STA, WTR, CT, and PE, as defined in Section~\ref{subsec:metrics}.

\subsection{Experiment 2: Unseen Instruction Generalisation}
\label{subsec:exp2}

Experiment~2 evaluates whether SGNav can generalise to unseen instruction formulations without retraining or manual semantic remapping. The target objects and environment remain the same as in Experiment~1, while only the language input is varied. As shown in Table~\ref{tab:exp2_instructions}, five instruction types are tested for each task, covering seen commands, synonym substitution, attribute-enriched descriptions, context-aware descriptions, and longer natural-language instructions.
\begin{table*}[t]
\centering
\caption{Instruction inputs used in Experiment~2.}
\label{tab:exp2_instructions}
\renewcommand{\arraystretch}{1.08}
\small
\begin{tabularx}{\textwidth}{p{0.15\textwidth}XXX}
\toprule
Type & Task~1: Magenta floating marker & Task~2: Large green buoy & Task~3: Yellow patrol boat \\
\midrule

Seen
& Go to the magenta target point.
& Go to the larger green buoy.
& Navigate to the yellow patrol boat. \\

Synonym
& Head to the pink-purple target point.
& Go to the emerald buoy.
& Head to the yellow maritime patrol boat. \\

Attribute
& Go to the bright magenta point marker.
& Go to the larger green buoy floating on the water.
& Go to the bright yellow patrol boat. \\

Context
& Ignore the boats and move to the magenta point.
& Go to the largest green buoy instead of the small one.
& Move toward the yellow vessel, not the magenta point. \\

Long
& Please move forward through the scene and stop when you reach the magenta target point.
& Please navigate across the water and stop at the largest green buoy.
& I want you to approach the yellow patrol boat as the final destination in this scene. \\

\bottomrule
\end{tabularx}
\end{table*}
The evaluation reports GA and SR, as defined in Section~\ref{subsec:metrics}. GA measures whether the instruction is grounded to the correct visual target, while SR measures whether the grounded target can be translated into successful closed-loop navigation.

\subsection{Experiment 3: Semantic Distractor Robustness}
Experiment~3 evaluates whether SGNav can identify the instructed target under distractor-rich conditions. For each task, the environment contains one target and three controlled distractors, designed to introduce colour-level, shape-level, and semantic-level ambiguity. As summarised in Table~\ref{tab:exp3_distractors_summary}, each distractor changes one dominant cue relative to the target, allowing the experiment to test whether SGNav grounds the complete semantic description rather than relying on a single visual attribute.

The evaluation reports CTR, DRR, WTR, and ColR, as defined in Section~\ref{subsec:metrics}. These metrics jointly evaluate correct target selection, distractor rejection, wrong-target failures, and navigation safety.
\begin{table*}[t]
\centering
\caption{Controlled distractor settings used in Experiment~3. Each task includes one target and three distractors corresponding to colour-level, shape-level, and semantic-level ambiguity.}
\label{tab:exp3_distractors_summary}
\renewcommand{\arraystretch}{1.05}
\setlength{\tabcolsep}{3pt}
\small
\begin{tabular}{p{0.09\textwidth} p{0.10\textwidth} p{0.22\textwidth} p{0.26\textwidth} p{0.22\textwidth}}
\toprule
Task & Target & Colour distractor & Shape distractor & Semantic distractor \\
\midrule

Task~1 &
Magenta floating marker &
Red floating marker
\newline
\includegraphics[width=0.16\textwidth]{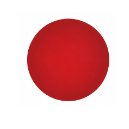} &
Magenta cylinder marker
\newline
\includegraphics[width=0.16\textwidth]{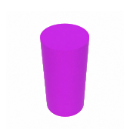} &
Magenta buoy
\newline
\includegraphics[width=0.16\textwidth]{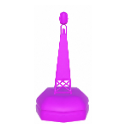} \\

Task~2 &
Large green buoy &
Large yellow buoy
\newline
\includegraphics[width=0.16\textwidth]{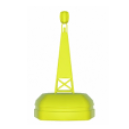} &
Large green marker
\newline
\includegraphics[width=0.16\textwidth]{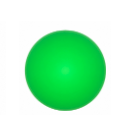} &
Small green buoy
\newline
\includegraphics[width=0.16\textwidth]{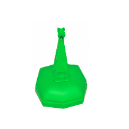} \\

Task~3 &
Yellow patrol boat &
Red patrol boat
\newline
\includegraphics[width=0.16\textwidth]{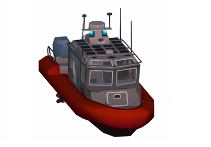} &
Yellow tugboat
\newline
\includegraphics[width=0.16\textwidth]{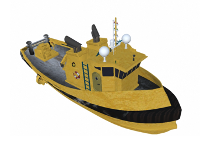} &
Yellow container ship
\newline
\includegraphics[width=0.16\textwidth]{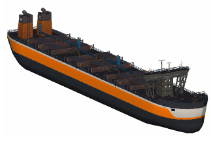} \\

\bottomrule
\end{tabular}
\end{table*}

\subsection{Experiment 4: Spatial Generalisation across Held-Out Layouts}
\label{subsec:exp4}

Experiment~4 evaluates whether SGNav can transfer from the training layout to held-out port layouts with controlled spatial variations. To isolate spatial generalisation from target-category variation, Task~1, magenta floating marker navigation, is used as the representative task. The policy is trained only in Layout~A and directly evaluated in Layouts~B--D without additional fine-tuning.

\begin{figure*}[t]
\centering
\includegraphics[width=0.8\textwidth]{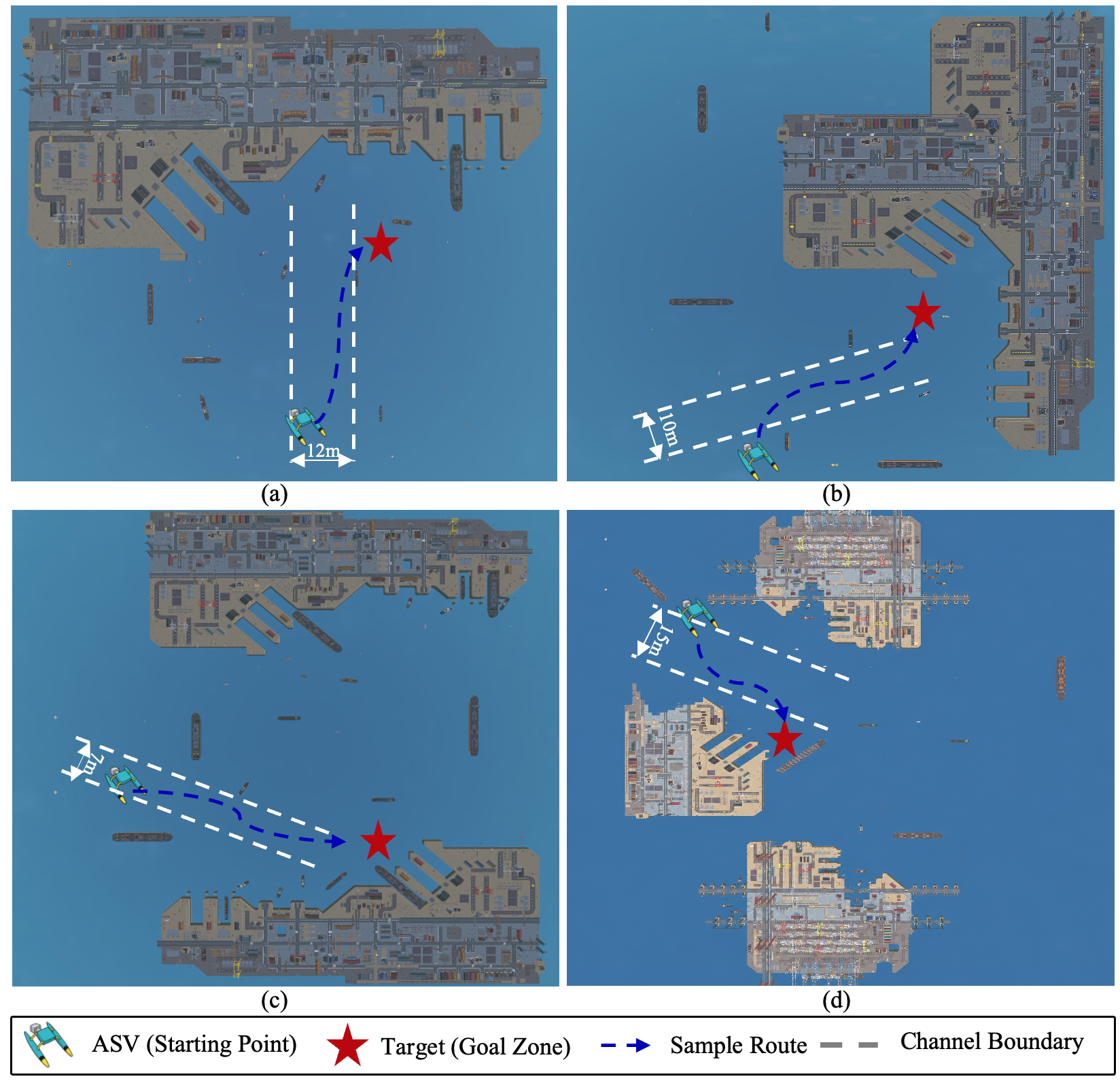}
\caption{
Controlled port layout variations used in Experiment~4. Panel~(a) shows the training layout, corresponding to Layout~A, while Panels~(b)--(d) show the held-out evaluation layouts, corresponding to Layouts~B--D. The four panels show the ASV starting point, target goal zone, sample route, and channel boundary. The layouts differ in corridor clearance, dock and pier configuration, obstacle distribution, and target placement.
}
\label{fig:exp4_layouts}
\end{figure*}

As shown in Fig.~\ref{fig:exp4_layouts}, the layout benchmark is constructed by varying interpretable spatial factors, including navigable corridor clearance, dock/pier topology, obstacle distribution, target placement, and the resulting feasible route. Layout~A serves as the training layout with a relatively wide and direct navigation corridor. Layout~B introduces a mild held-out variation with a narrower corridor and a changed approach route. Layout~C creates a more constrained setting with the narrowest corridor and higher obstacle density. Layout~D represents a different open-basin configuration with multiple port regions and a longer-range approach. This controlled layout design reduces the arbitrariness of the evaluation and allows the transfer performance to be analysed under different degrees of spatial shift.

Table~\ref{tab:exp4_layout_factors} summarises the controlled layout factors. The corridor clearance refers to the minimum usable lateral clearance along the feasible route from the ASV starting point to the target goal zone. Obstacle complexity is assigned qualitatively according to the number and proximity of non-target vessels, static structures, and navigation markers along the feasible route corridor.

\begin{table}[t]
\centering
\caption{Controlled layout factors used in Experiment~4.}
\label{tab:exp4_layout_factors}
\renewcommand{\arraystretch}{1.1}
\begin{tabular}{p{0.09\linewidth} p{0.12\linewidth} p{0.28\linewidth} p{0.14\linewidth} p{0.30\linewidth}}
\toprule
Layout & Corridor clearance & Dock/pier topology & Obstacle complexity & Target placement \\
\midrule
A & 12 m & Single-side straight dock & Low & Open channel, fully visible \\
B & 10 m & Right-side modified pier & Medium & Offset target, changed approach route \\
C & 7 m & Bilateral narrow passage & High & Near berth, constrained approach \\
D & 15 m & Multi-region open basin & Medium & Open basin, long-range approach \\
\bottomrule
\end{tabular}
\end{table}

SGNav is compared with the same baselines used in Experiment~1, including No-Target PPO, Vision-Only PPO, and Oracle PPO. The evaluation reports SR, STA, ColR, CT, and PE, as defined in Section~\ref{subsec:metrics}, to measure task completion, semantic grounding reliability, navigation safety, and trajectory efficiency under spatial layout changes.

\subsection{Experiment 5: Ablation Study}
\label{subsec:exp5}

Experiment~5 analyses the contribution of selected SGNav components after language-grounded target resolution has been established. Two representative tasks are selected: Task~1, magenta floating marker navigation, and Task~3, yellow patrol boat navigation. Task~1 mainly tests colour- and shape-based target grounding, while Task~3 focuses on vessel-category discrimination under stronger maritime object interference. Task~2 is omitted to avoid redundancy, as the ablation study focuses on these two complementary diagnostic settings.

The Full SGNav model is compared with two ablation variants: (i) without harbour-aware filtering and (ii) without semantic consistency. The necessity of language-grounded target resolution has already been evaluated through the Vision-Only PPO and No-Target PPO baselines in Experiment~1. Therefore, this ablation study focuses on whether distractor suppression and semantic consistency improve robust target selection and navigation after target grounding has been established. The evaluation reports SR, STA, WTR, and DRR, as defined in Section~\ref{subsec:metrics}, to measure task success, semantic target accuracy, wrong-target failures, and distractor rejection.
\FloatBarrier

\section{Experiment Results}
\label{sec:5}

\subsection{Results of Experiment 1} 
The quantitative results of Experiment~1 are reported in Table~\ref{tab:exp1_results}. SGNav achieves consistently high success rates across the three semantic target navigation tasks, with SR values of $97.0\pm1.2\%$, $92.0\pm1.5\%$, and $90.0\pm1.8\%$ for Tasks~1--3, respectively. These results are close to the Oracle PPO upper bound and substantially higher than the Vision-Only PPO and No-Target PPO baselines, indicating that language-grounded target resolution is essential for target-specific maritime navigation.

The performance gap between SGNav and the non-semantic baselines is particularly clear across all tasks. Vision-Only PPO and No-Target PPO achieve only limited success in Tasks~1 and~2 and fail to complete Task~3, showing that visual observations or environment states alone are insufficient for reliable target-specific navigation in cluttered harbour scenes. In contrast, SGNav maintains high semantic target accuracy across all tasks, with STA values above $97\%$ and WTR values below $3\%$. This suggests that SGNav usually reaches the intended semantic target rather than succeeding through accidental arrival or visually driven exploration.

Compared with Oracle PPO, SGNav achieves similar SR but generally requires longer completion time and shows slightly lower path efficiency. This is expected because Oracle PPO receives privileged ground-truth target information, whereas SGNav must infer the target from language-grounded visual perception. The results therefore show that the proposed semantic grounding module can provide reliable target information for downstream navigation, while still leaving room for improvement in trajectory efficiency.

\begin{table*}[t]
\centering
\caption{Performance comparison of different methods in Experiment~1. Results are reported as mean $\pm$ standard deviation over three random seeds.}
\label{tab:exp1_results}
\renewcommand{\arraystretch}{1.12}
\setlength{\tabcolsep}{3.5pt}
\small
\begin{tabular}{llccccc}
\toprule
Task & Method
& SR (\%)
& STA (\%)
& WTR (\%)
& CT (sim. s)
& PE \\
\midrule

\multirow{4}{*}{Task~1}
& Vision-Only PPO
& $10.3 \pm 1.5$
& N/A
& N/A
& $489.9 \pm 18.6$
& $0.745 \pm 0.031$ \\
& No-Target PPO
& $5.2 \pm 1.0$
& N/A
& N/A
& $669.1 \pm 24.7$
& $0.423 \pm 0.028$ \\
& Oracle PPO
& $97.0 \pm 1.5$
& N/A
& N/A
& $157.6 \pm 7.9$
& $0.996 \pm 0.012$ \\
& \textbf{Proposed SGNav}
& $\mathbf{97.0 \pm 1.2}$
& $\mathbf{98.59 \pm 0.64}$
& $\mathbf{1.41 \pm 0.64}$
& $\mathbf{283.0 \pm 12.4}$
& $\mathbf{0.821 \pm 0.027}$ \\

\midrule

\multirow{4}{*}{Task~2}
& Vision-Only PPO
& $4.1 \pm 0.9$
& N/A
& N/A
& $709.6 \pm 31.5$
& $0.435 \pm 0.024$ \\
& No-Target PPO
& $1.0 \pm 0.6$
& N/A
& N/A
& $869.1 \pm 36.8$
& $0.223 \pm 0.019$ \\
& Oracle PPO
& $92.0 \pm 1.7$
& N/A
& N/A
& $384.6 \pm 16.3$
& $0.877 \pm 0.022$ \\
& \textbf{Proposed SGNav}
& $\mathbf{92.0 \pm 1.5}$
& $\mathbf{97.09 \pm 0.81}$
& $\mathbf{2.91 \pm 0.81}$
& $\mathbf{400.7 \pm 17.6}$
& $\mathbf{0.799 \pm 0.025}$ \\

\midrule

\multirow{4}{*}{Task~3}
& Vision-Only PPO
& $0.0 \pm 0.0$
& N/A
& N/A
& N/A
& N/A \\
& No-Target PPO
& $0.0 \pm 0.0$
& N/A
& N/A
& N/A
& N/A \\
& Oracle PPO
& $90.0 \pm 2.0$
& N/A
& N/A
& $466.3 \pm 19.8$
& $0.894 \pm 0.027$ \\
& \textbf{Proposed SGNav}
& $\mathbf{90.0 \pm 1.8}$
& $\mathbf{97.65 \pm 0.72}$
& $\mathbf{2.35 \pm 0.72}$
& $\mathbf{502.8 \pm 21.4}$
& $\mathbf{0.830 \pm 0.030}$ \\

\bottomrule
\end{tabular}

\vspace{0.5em}
\begin{minipage}{0.96\textwidth}
\footnotesize
\textit{Note:} N/A indicates that the metric is not applicable. STA and WTR are not reported for No-Target PPO, Vision-Only PPO, and Oracle PPO because these baselines do not perform explicit language-grounded semantic target resolution. CT and PE are reported only for successful episodes and are therefore N/A when no successful episode is achieved. CT denotes simulated episode completion time rather than wall-clock training or inference time.
\end{minipage}
\end{table*}

\begin{figure}[H] 
\centering \includegraphics[width=\textwidth]{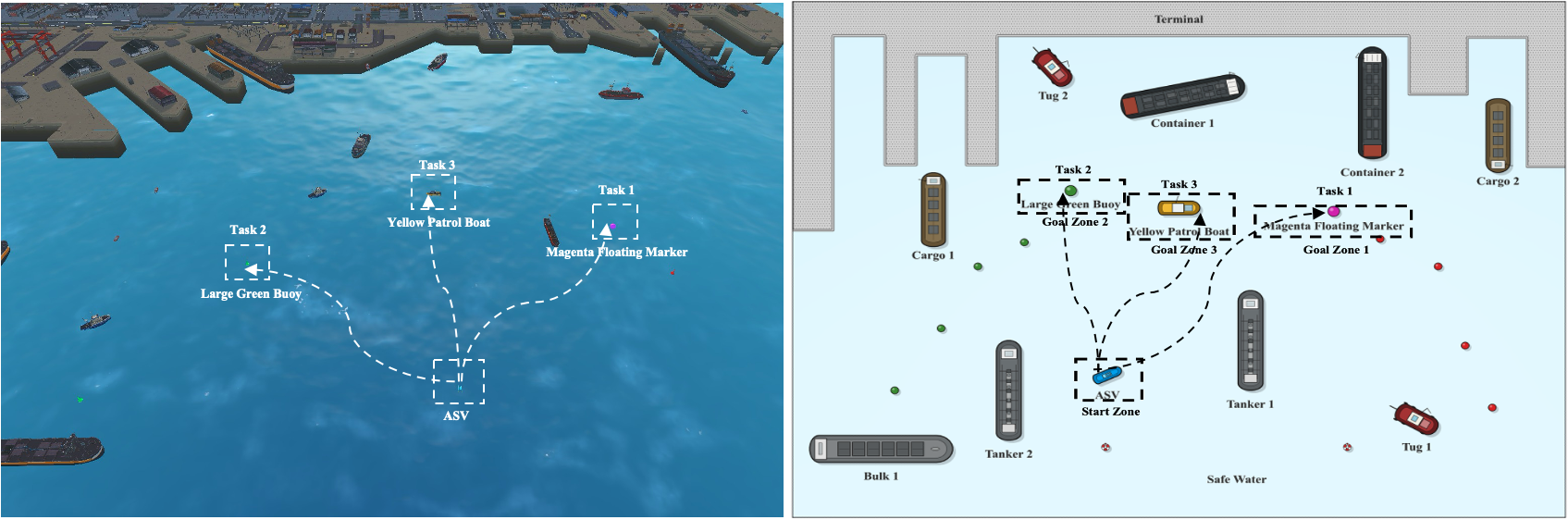} 
\caption{ Representative navigation trajectories for the three semantic navigation tasks in Experiment~1. From top to bottom, the rows correspond to magenta floating marker, large green buoy, and yellow patrol boat navigation. Each row shows the simulator view and the corresponding simplified top-down layout, where the dashed trajectory illustrates the ASV motion from the shared initial position toward the instructed target. } 
\label{fig:exp1_123traj}
\end{figure}

\begin{figure*}[t]
\centering

\begin{minipage}[t]{0.31\textwidth}
    \centering
    \includegraphics[width=\linewidth]{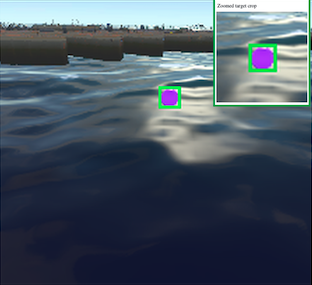}
    
    \vspace{0.2em}
    {\small \textbf{(a) Success 1: early grounding}\\
    Target visible.}
\end{minipage}
\hfill
\begin{minipage}[t]{0.31\textwidth}
    \centering
    \includegraphics[width=\linewidth]{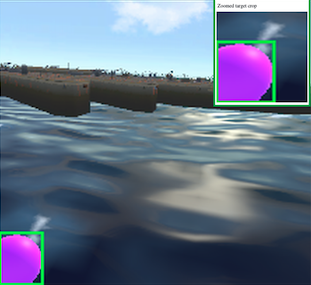}
    
    \vspace{0.2em}
    {\small \textbf{(b) Success 2: target boxed}\\
    DINO target box.}
\end{minipage}
\hfill
\begin{minipage}[t]{0.31\textwidth}
    \centering
    \includegraphics[width=\linewidth]{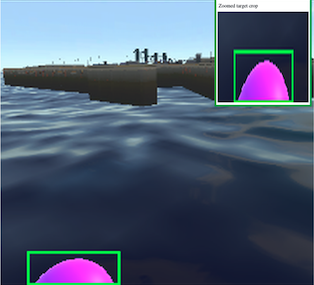}
    
    \vspace{0.2em}
    {\small \textbf{(c) Success 3: repeated case}\\
    Target re-detected.}
\end{minipage}

\vspace{0.8em}

\begin{minipage}[t]{0.31\textwidth}
    \centering
    \includegraphics[width=\linewidth]{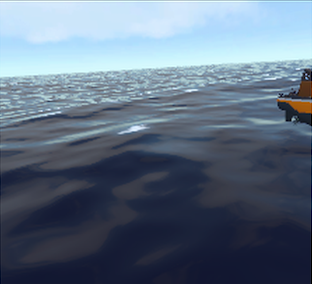}
    
    \vspace{0.2em}
    {\small \textbf{(d) Failure 1: out of bounds}\\
    Trajectory leaves valid area after grounding.}
\end{minipage}
\hfill
\begin{minipage}[t]{0.31\textwidth}
    \centering
    \includegraphics[width=\linewidth]{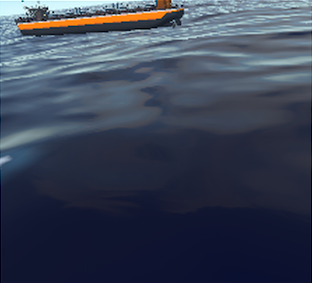}
    
    \vspace{0.2em}
    {\small \textbf{(e) Failure 2: timeout}\\
    Target not visible in final frame.}
\end{minipage}
\hfill
\begin{minipage}[t]{0.31\textwidth}
    \centering
    \includegraphics[width=\linewidth]{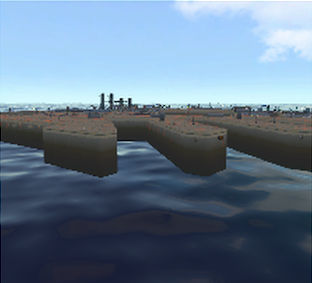}
    
    \vspace{0.2em}
    {\small \textbf{(f) Failure 3: obstacle}\\
    Target lost near harbour structures.}
\end{minipage}

\caption{Representative success and failure episodes for Task 1: Magenta Floating Marker navigation. The top row shows representative successful cases, where the target is detected from different viewpoints and distances. The bottom row shows failure cases, including out-of-bounds and timeout cases, which indicate that navigation can still fail when the ASV leaves the valid area or cannot complete the approach within the step limit. } 
\label{fig:task1_grounding} 
\end{figure*}

\begin{figure*}[t]
\centering
\begin{minipage}[t]{0.31\textwidth}
    \centering
    \includegraphics[width=\linewidth]{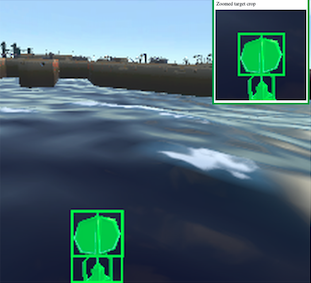}
    
    \vspace{0.2em}
    {\small \textbf{(a) Success 1: buoy grounded}\\
    Larger green buoy detected.}
\end{minipage}
\hfill
\begin{minipage}[t]{0.31\textwidth}
    \centering
    \includegraphics[width=\linewidth]{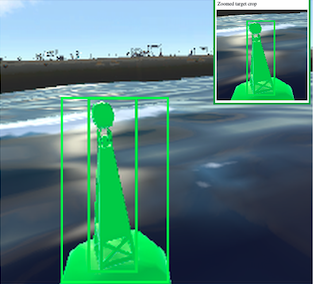}
    
    \vspace{0.2em}
    {\small \textbf{(b) Success 2: stable tracking}\\
    Target remains visible.}
\end{minipage}
\hfill
\begin{minipage}[t]{0.31\textwidth}
    \centering
    \includegraphics[width=\linewidth]{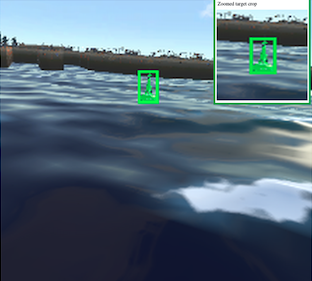}
    
    \vspace{0.2em}
    {\small \textbf{(c) Success 3: approach}\\
    Target-guided approach.}
\end{minipage}

\vspace{0.8em}

\begin{minipage}[t]{0.31\textwidth}
    \centering
    \includegraphics[width=\linewidth]{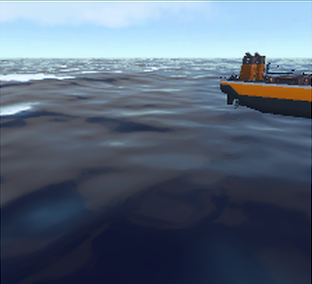}
    
    \vspace{0.2em}
    {\small \textbf{(d) Failure 1: out of bounds}\\
    Trajectory leaves valid area.}
\end{minipage}
\hfill
\begin{minipage}[t]{0.31\textwidth}
    \centering
    \includegraphics[width=\linewidth]{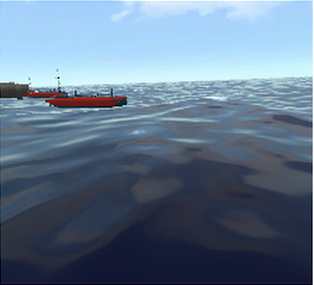}
    
    \vspace{0.2em}
    {\small \textbf{(e) Failure 2: obstacle}\\
    Local collision failure.}
\end{minipage}
\hfill
\begin{minipage}[t]{0.31\textwidth}
    \centering
    \includegraphics[width=\linewidth]{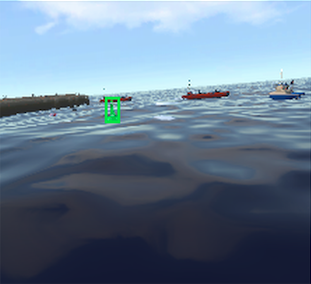}
    
    \vspace{0.2em}
    {\small \textbf{(f) Failure 3: timeout}\\
    No completion within step limit.}
\end{minipage}

\caption{Representative success and failure episodes for Task 2: Green Buoy navigation. Successful cases demonstrate that SGNav can ground the buoy, maintain stable tracking, and guide the ASV toward the target. Failure cases include out-of-bounds, obstacle-related failure, and timeout, showing the influence of navigation constraints and surrounding distractors on closed-loop execution. } \label{fig:task2_grounding} 
\end{figure*}

\begin{figure*}[t]
\centering

\begin{minipage}[t]{0.31\textwidth}
    \centering
    \includegraphics[width=\linewidth]{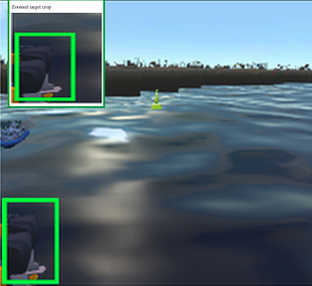}
    
    \vspace{0.2em}
    {\small \textbf{(a) Success 1: target visible}\\
    Patrol boat detected.}
\end{minipage}
\hfill
\begin{minipage}[t]{0.31\textwidth}
    \centering
    \includegraphics[width=\linewidth]{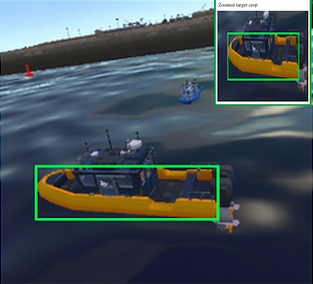}
    
    \vspace{0.2em}
    {\small \textbf{(b) Success 2: close grounding}\\
    High-confidence target cue.}
\end{minipage}
\hfill
\begin{minipage}[t]{0.31\textwidth}
    \centering
    \includegraphics[width=\linewidth]{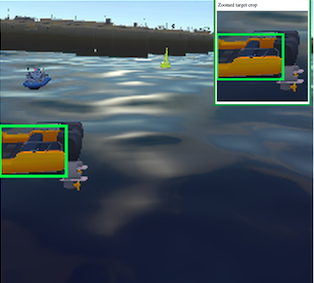}
    
    \vspace{0.2em}
    {\small \textbf{(c) Success 3: repeated cue}\\
    Target remains localized.}
\end{minipage}

\vspace{0.8em}

\begin{minipage}[t]{0.31\textwidth}
    \centering
    \includegraphics[width=\linewidth]{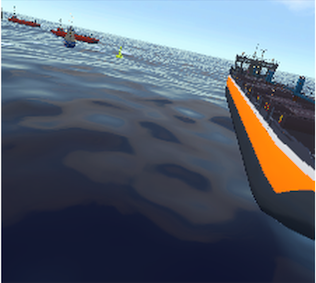}
    
    \vspace{0.2em}
    {\small \textbf{(d) Failure 1: obstacle}\\
    Collision despite target grounding.}
\end{minipage}
\hfill
\begin{minipage}[t]{0.31\textwidth}
    \centering
    \includegraphics[width=\linewidth]{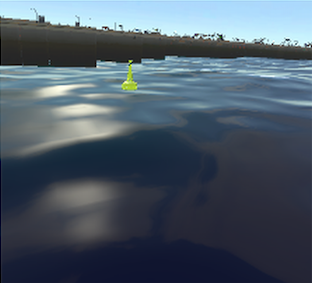}
    
    \vspace{0.2em}
    {\small \textbf{(e) Failure 2: out of bounds}\\
    Trajectory leaves valid area.}
\end{minipage}
\hfill
\begin{minipage}[t]{0.31\textwidth}
    \centering
    \includegraphics[width=\linewidth]{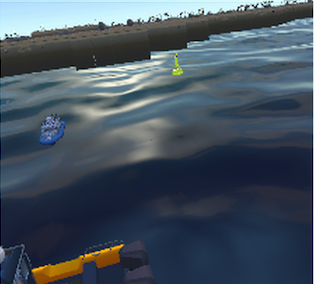}
    
    \vspace{0.2em}
    {\small \textbf{(f) Failure 3: timeout}\\
    No completion within step limit.}
\end{minipage}

\caption{Representative success and failure episodes for Task 3: Yellow Patrol Boat navigation. Successful cases show that the target Patrol Boat can be localized when it is visible, closely grounded during approach, and repeatedly detected across frames. Failure cases include obstacle collision, out-of-bounds, and timeout, suggesting that large maritime targets can be grounded reliably but still introduce navigation challenges in cluttered scenes. } 
\label{fig:task3_grounding} 
\end{figure*}

Figure~\ref{fig:exp1_123traj} further visualizes the navigation behaviors in the three tasks. Starting from the same ASV position, the generated trajectories move toward the instructed semantic targets, namely the Magenta Floating Marker, Green Buoy, and Yellow Patrol Boat. This confirms that the navigation policy is conditioned on the target instruction rather than following a fixed route. The grounding examples in Figs.~\ref{fig:task1_grounding}, \ref{fig:task2_grounding}, and \ref{fig:task3_grounding} provide additional qualitative evidence. In successful cases, SGNav can detect the target at different distances and viewpoints, maintain target localization during approach, and repeatedly recover the same semantic cue across frames. The failure cases reveal the main limitations of the current framework, including leaving the valid navigation area, failing to complete the task within the step limit, and obstacle-related failures. These results suggest that while SGNav provides reliable semantic grounding and target-conditioned navigation, its closed-loop control still remains sensitive to boundary constraints, obstacle configurations, and long-horizon navigation errors.

\subsection{Results of Experiment 2}

\begin{figure*}[t]
\centering
\includegraphics[width=\textwidth]{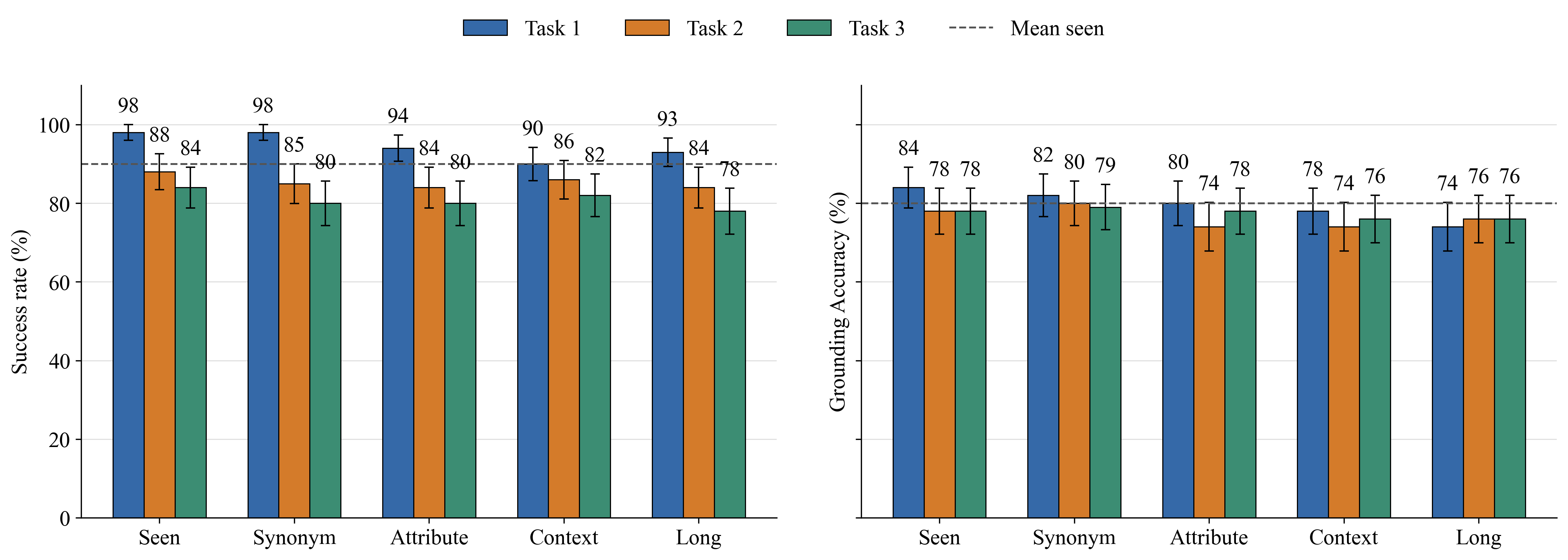}
\caption{
Instruction generalisation results in Experiment~2. The left panel reports the success rate under five instruction types, including seen instructions, synonym substitution, attribute-enriched descriptions, context-aware descriptions, and long natural-language instructions. The right panel reports the corresponding grounding accuracy. Results are shown for Tasks~1--3. Error bars indicate the standard deviation over repeated evaluation runs, and the dashed horizontal line indicates the mean performance under seen instructions.
}
\label{fig:exp2_instruction_generalisation}
\end{figure*}

Figure~\ref{fig:exp2_instruction_generalisation} summarises the instruction generalisation results. SGNav remains robust when the input changes from seen instructions to unseen linguistic variants. The mean SR under seen instructions is about 90\%, while the unseen categories remain close to this reference, with synonym substitution around 87.7\% and the other instruction types around 85--86\%. GA is slightly more sensitive to instruction complexity: the mean GA under seen instructions is about 80\%, and the unseen categories generally remain above 75\%, with synonym substitution closest to the seen reference. The decrease under attribute-enriched, context-aware, and long instructions suggests that additional linguistic content affects visual grounding more directly than final navigation success. Therefore, SGNav is not limited to fixed command templates and can generalise to alternative wording, added attributes, contextual cues, and longer natural-language commands.

\begin{figure*}[t]
\centering

\begin{minipage}[t]{0.31\textwidth}
    \centering
    \includegraphics[width=\linewidth]{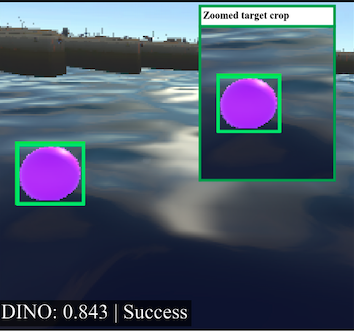}
    
    \vspace{0.2em}
    {\small \textbf{(a) Seen}\\
    Go to the magenta target point.}
\end{minipage}
\hfill
\begin{minipage}[t]{0.31\textwidth}
    \centering
    \includegraphics[width=\linewidth]{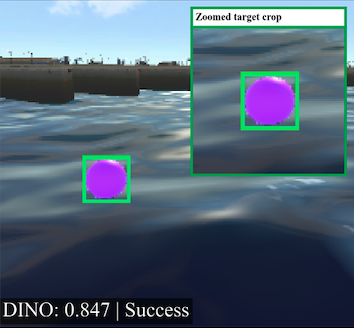}
    
    \vspace{0.2em}
    {\small \textbf{(b) Synonym}\\
    Head to the pink-purple target point.}
\end{minipage}
\hfill
\begin{minipage}[t]{0.31\textwidth}
    \centering
    \includegraphics[width=\linewidth]{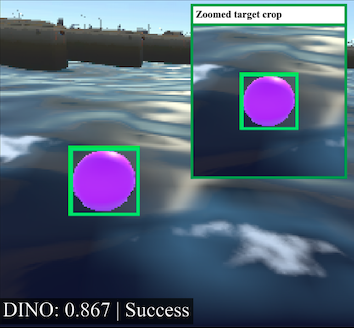}
    
    \vspace{0.2em}
    {\small \textbf{(c) Attribute}\\
    Go to the bright magenta point marker.}
\end{minipage}

\vspace{0.8em}

\makebox[\textwidth][c]{%
\begin{minipage}[t]{0.31\textwidth}
    \centering
    \includegraphics[width=\linewidth]{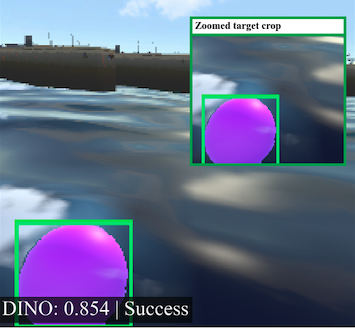}
    
    \vspace{0.2em}
    {\small \textbf{(d) Context}\\
    Ignore the boats and move to the magenta point.}
\end{minipage}
\hspace{0.06\textwidth}
\begin{minipage}[t]{0.31\textwidth}
    \centering
    \includegraphics[width=\linewidth]{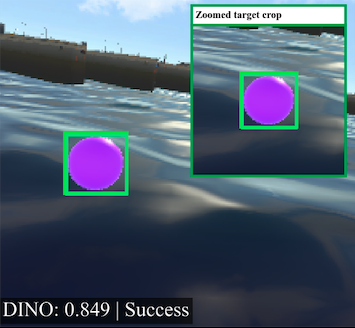}
    
    \vspace{0.2em}
    {\small \textbf{(e) Long}\\
    Please move forward through the scene and stop when you reach the magenta target point.}
\end{minipage}
}

\caption{
Qualitative results of Task~1 under different semantic instruction types. The target is the magenta point object, and the five cases correspond to seen instruction, synonym substitution, attribute-enriched description, context-aware description, and long natural-language instruction.
}
\label{fig:task1_instruction_variations}
\end{figure*}

\begin{figure*}[t]
\centering

\begin{minipage}[t]{0.31\textwidth}
    \centering
    \includegraphics[width=\linewidth]{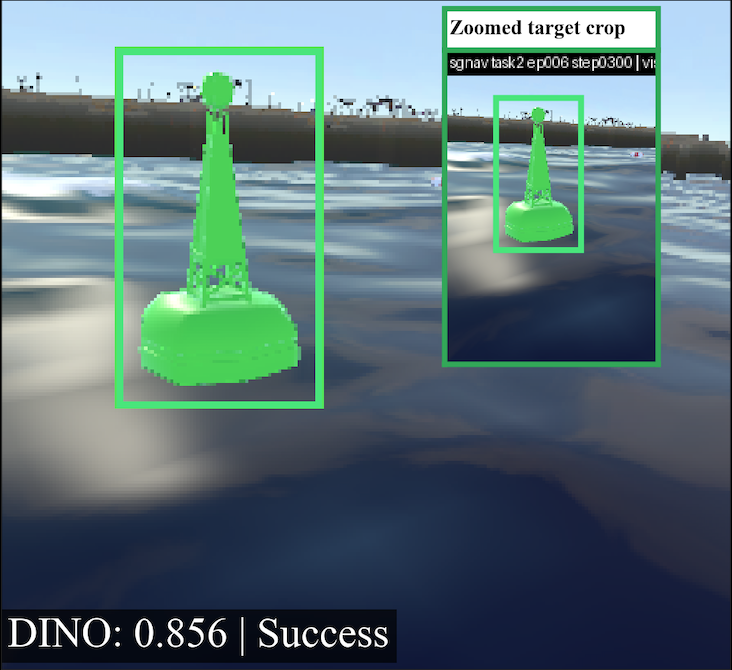}

    \vspace{0.2em}
    {\small \textbf{(a) Seen}\\
    Go to the larger green buoy.}
\end{minipage}
\hfill
\begin{minipage}[t]{0.31\textwidth}
    \centering
    \includegraphics[width=\linewidth]{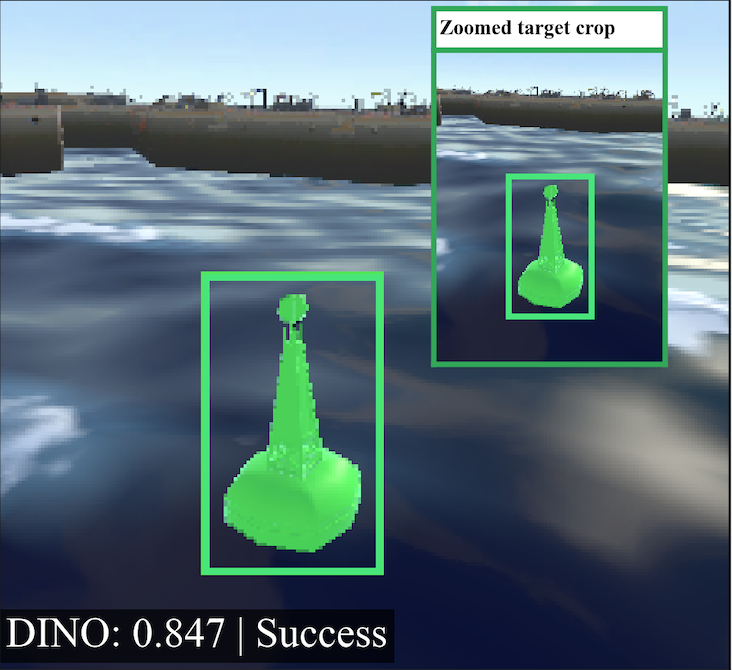}

    \vspace{0.2em}
    {\small \textbf{(b) Synonym}\\
    Go to the emerald buoy.}
\end{minipage}
\hfill
\begin{minipage}[t]{0.31\textwidth}
    \centering
    \includegraphics[width=\linewidth]{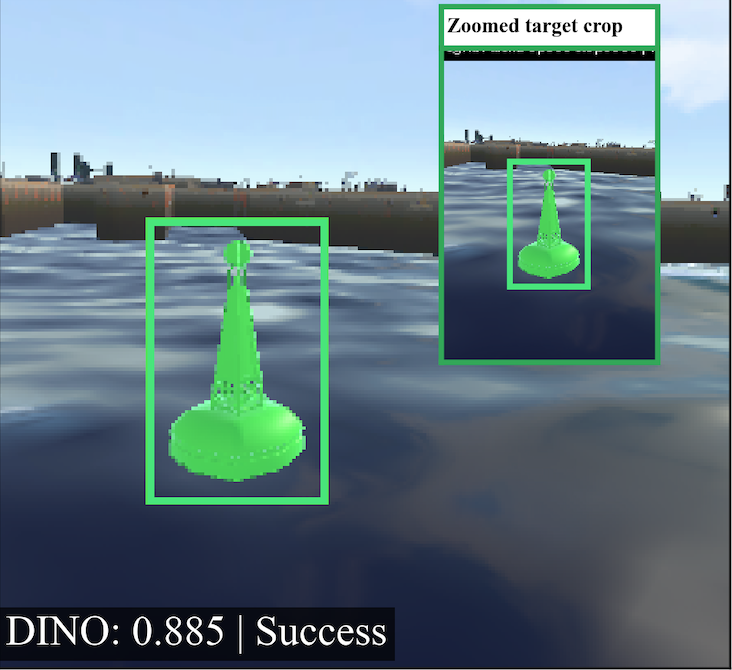}

    \vspace{0.2em}
    {\small \textbf{(c) Attribute}\\
    Go to the larger green buoy floating on the water.}
\end{minipage}

\vspace{0.8em}

\makebox[\textwidth][c]{%
\begin{minipage}[t]{0.31\textwidth}
    \centering
    \includegraphics[width=\linewidth]{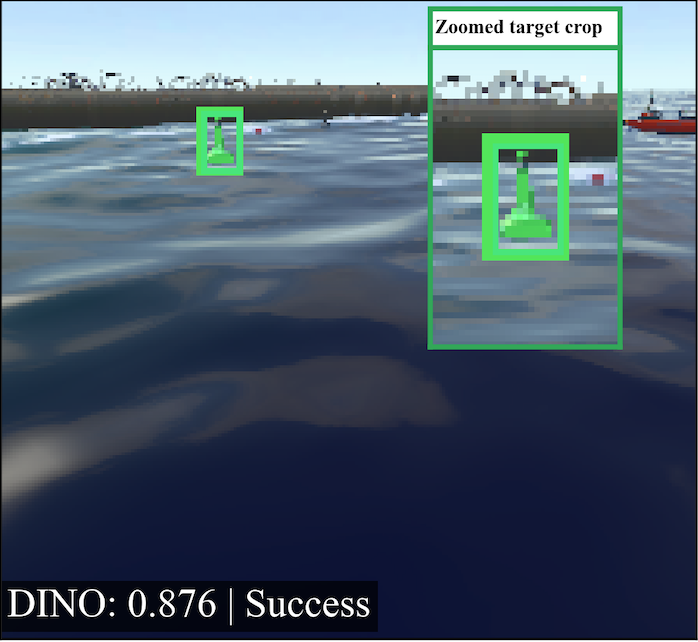}

    \vspace{0.2em}
    {\small \textbf{(d) Context}\\
    Go to the largest green buoy instead of the small one.}
\end{minipage}
\hspace{0.06\textwidth}
\begin{minipage}[t]{0.31\textwidth}
    \centering
    \includegraphics[width=\linewidth]{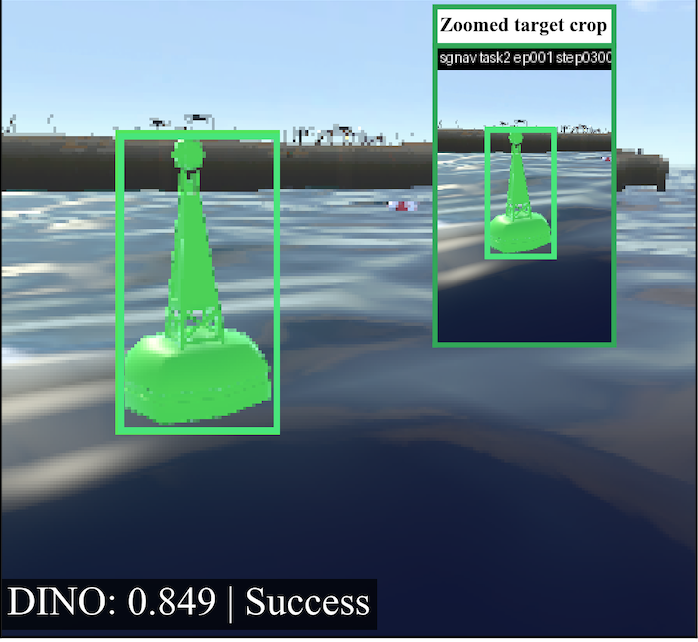}

    \vspace{0.2em}
    {\small \textbf{(e) Long}\\
    Please navigate across the water and stop at the largest green buoy.}
\end{minipage}
}

\caption{
Qualitative results of Task~2 under different semantic instruction types. The target is the large green buoy, and the five cases correspond to seen instruction, synonym substitution, attribute-enriched description, context-aware description, and long natural-language instruction.
}
\label{fig:task2_instruction_variations}
\end{figure*}

\begin{figure*}[t]
\centering

\begin{minipage}[t]{0.31\textwidth}
    \centering
    \includegraphics[width=\linewidth]{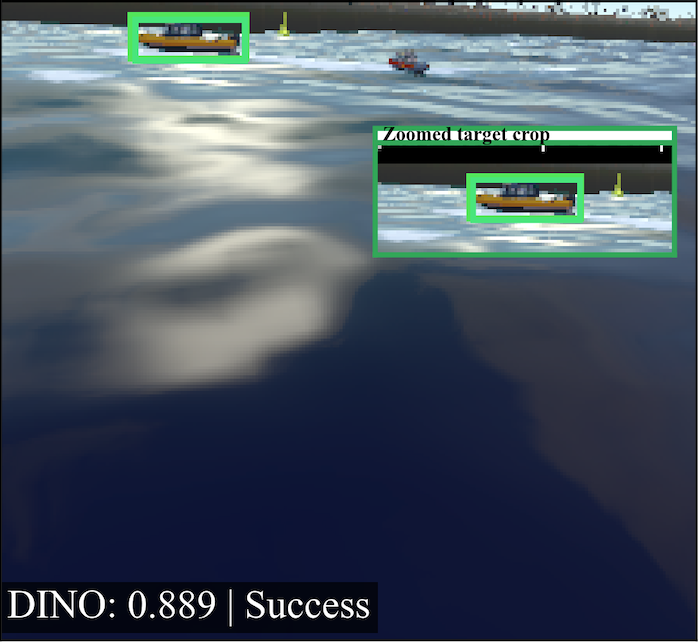}

    \vspace{0.2em}
    {\small \textbf{(a) Seen}\\
    Navigate to the yellow patrol boat.}
\end{minipage}
\hfill
\begin{minipage}[t]{0.31\textwidth}
    \centering
    \includegraphics[width=\linewidth]{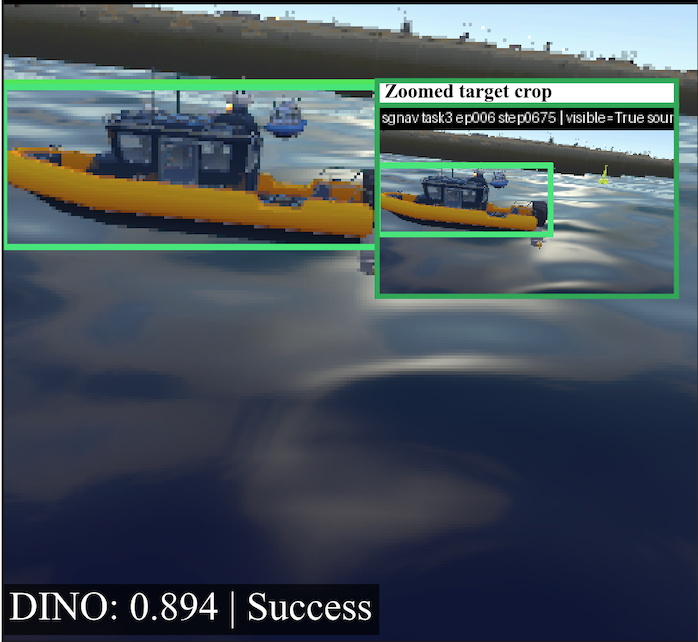}

    \vspace{0.2em}
    {\small \textbf{(b) Synonym}\\
    Head to the yellow maritime patrol boat.}
\end{minipage}
\hfill
\begin{minipage}[t]{0.31\textwidth}
    \centering
    \includegraphics[width=\linewidth]{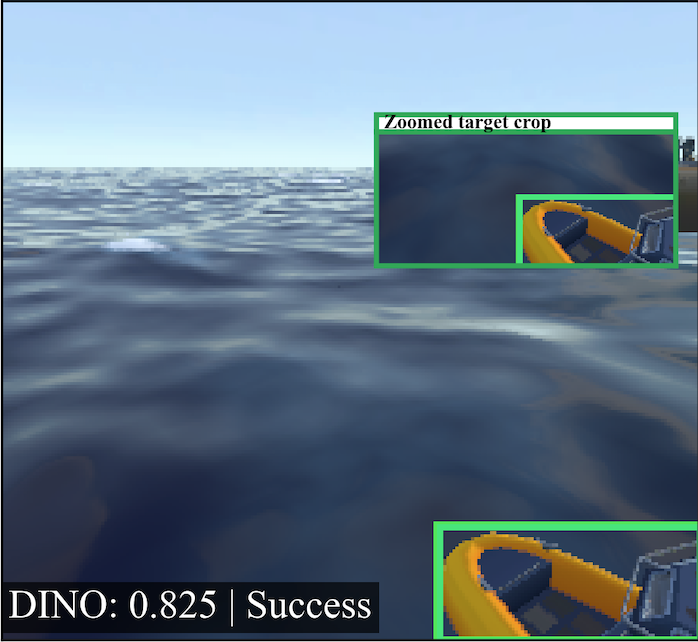}

    \vspace{0.2em}
    {\small \textbf{(c) Attribute}\\
    Go to the bright yellow patrol boat.}
\end{minipage}

\vspace{0.8em}

\makebox[\textwidth][c]{%
\begin{minipage}[t]{0.31\textwidth}
    \centering
    \includegraphics[width=\linewidth]{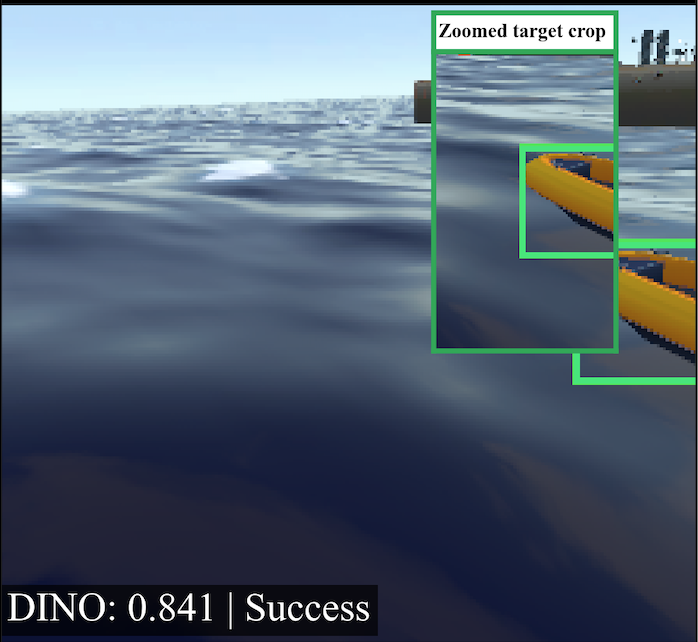}

    \vspace{0.2em}
    {\small \textbf{(d) Context}\\
    Move toward the yellow vessel, not the magenta point.}
\end{minipage}
\hspace{0.06\textwidth}
\begin{minipage}[t]{0.31\textwidth}
    \centering
    \includegraphics[width=\linewidth]{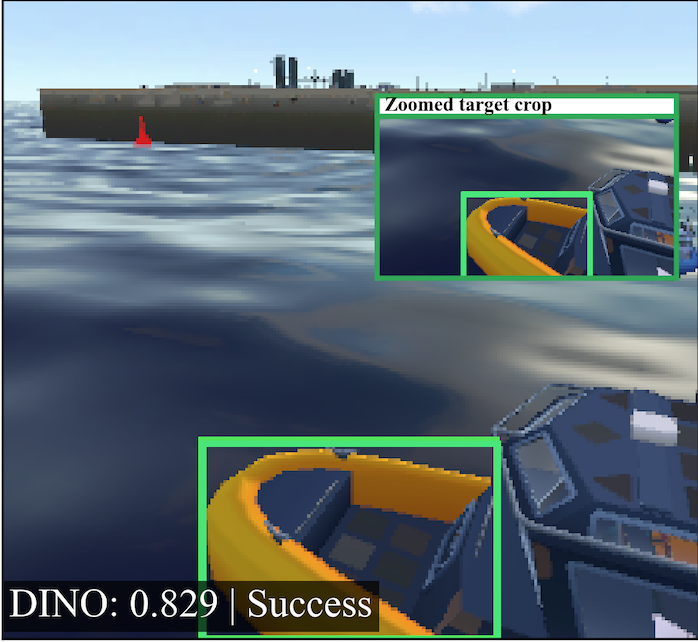}

    \vspace{0.2em}
    {\small \textbf{(e) Long}\\
    I want you to approach the yellow patrol boat as the final destination in this scene.}
\end{minipage}
}

\caption{
Qualitative results of Task~3 under different semantic instruction types. The target is the yellow patrol boat, and the five cases correspond to seen instruction, synonym substitution, attribute-enriched description, context-aware description, and long natural-language instruction.
}
\label{fig:task3_instruction_variations}
\end{figure*}

A clear task-level trend can also be observed. Task~1 achieves the best performance because the magenta target point is visually simple and distinctive. Task~2 shows a moderate decrease due to the more structured buoy appearance and the presence of buoy-like objects. Task~3 is the most challenging, as the yellow patrol boat is a larger vessel with viewpoint-dependent and partially visible appearances. The qualitative examples in Figs.~\ref{fig:task1_instruction_variations}--\ref{fig:task3_instruction_variations}, selected from the episodes with the highest GroundingDINO confidence for each instruction type, further show that SGNav can localize the target under different expressions, while complex vessel targets lead to more unstable visual grounding. Hence, the horizontal comparison shows that synonym and attribute variations are easier to generalise, whereas context-aware and long commands are more difficult. The vertical comparison shows a gradual performance decrease from Task~1 to Task~3, reflecting increasing visual and navigational complexity.

\subsection{Results of Experiment 3}

Table~\ref{tab:exp3_distractor} reports the robustness of SGNav under controlled distractor settings. The table separates task-level navigation outcomes, including CTR, WTR, and ColR, from distractor-type-specific rejection performance, measured by $\mathrm{DRR}_{\mathrm{colour}}$, $\mathrm{DRR}_{\mathrm{shape}}$, and $\mathrm{DRR}_{\mathrm{semantic}}$. Overall, SGNav maintains high robustness across all three tasks, with CTR remaining above $96\%$ and ColR staying below $1.0\%$ on average. This indicates that the proposed framework can usually select the instructed target and complete navigation safely even when visually or semantically related distractors are present.

A clearer difference appears in the distractor rejection results. For Task~1, SGNav achieves high rejection rates for colour and shape distractors, with $\mathrm{DRR}_{\mathrm{colour}}=99.0\pm1.0\%$ and $\mathrm{DRR}_{\mathrm{shape}}=96.3\pm2.1\%$, but the semantic distractor is more challenging, reducing $\mathrm{DRR}_{\mathrm{semantic}}$ to $91.7\pm3.2\%$. This suggests that the magenta buoy-like distractor provides a stronger semantic competitor to the magenta floating marker than simple colour or shape variations. For Tasks~2 and~3, the rejection rates remain consistently high across all distractor types, showing that the large green buoy and yellow patrol boat provide richer visual and semantic cues for target discrimination.

The wrong-target and collision rates remain low across the three tasks, with WTR values of $2.3\pm1.5\%$, $3.0\pm1.7\%$, and $1.7\pm1.5\%$ for Tasks~1--3, respectively. These results show that semantic distractors are generally more challenging than simple colour or shape distractors, but SGNav still preserves strong target selection accuracy, distractor rejection, and navigation safety under distractor-rich conditions.

\begin{table}[t]
\centering
\caption{Performance of SGNav under controlled distractor settings in Experiment 3. Results are reported as mean $\pm$ standard deviation over three random seeds.}
\label{tab:exp3_distractor}
\renewcommand{\arraystretch}{1.12}
\setlength{\tabcolsep}{4pt}
\small
\begin{tabular}{lcccccc}
\toprule
Task 
& CTR (\%) 
& DRR$_{\mathrm{colour}}$ (\%) 
& DRR$_{\mathrm{shape}}$ (\%) 
& DRR$_{\mathrm{semantic}}$ (\%) 
& WTR (\%) 
& ColR (\%) \\
\midrule

Task~1 
& $97.7 \pm 1.5$
& $99.0 \pm 1.0$
& $96.3 \pm 2.1$
& $91.7 \pm 3.2$
& $2.3 \pm 1.5$
& $0.3 \pm 0.6$ \\

Task~2 
& $96.0 \pm 2.0$
& $96.7 \pm 1.5$
& $98.7 \pm 1.5$
& $95.3 \pm 2.1$
& $3.0 \pm 1.7$
& $1.0 \pm 1.0$ \\

Task~3 
& $98.0 \pm 1.7$
& $99.3 \pm 1.2$
& $98.0 \pm 1.7$
& $97.3 \pm 1.5$
& $1.7 \pm 1.5$
& $0.7 \pm 0.6$ \\

\bottomrule
\end{tabular}
\end{table}

\begin{figure*}[t]
\centering
\setlength{\tabcolsep}{3pt}
\renewcommand{\arraystretch}{1.15}
\small
\begin{tabular}{c c c c}
\toprule
 & \textbf{Colour distractor} & \textbf{Shape distractor} & \textbf{Semantic distractor} \\
\midrule

\textbf{Task~1} &
\begin{minipage}{0.28\textwidth}
\centering
\includegraphics[width=\linewidth]{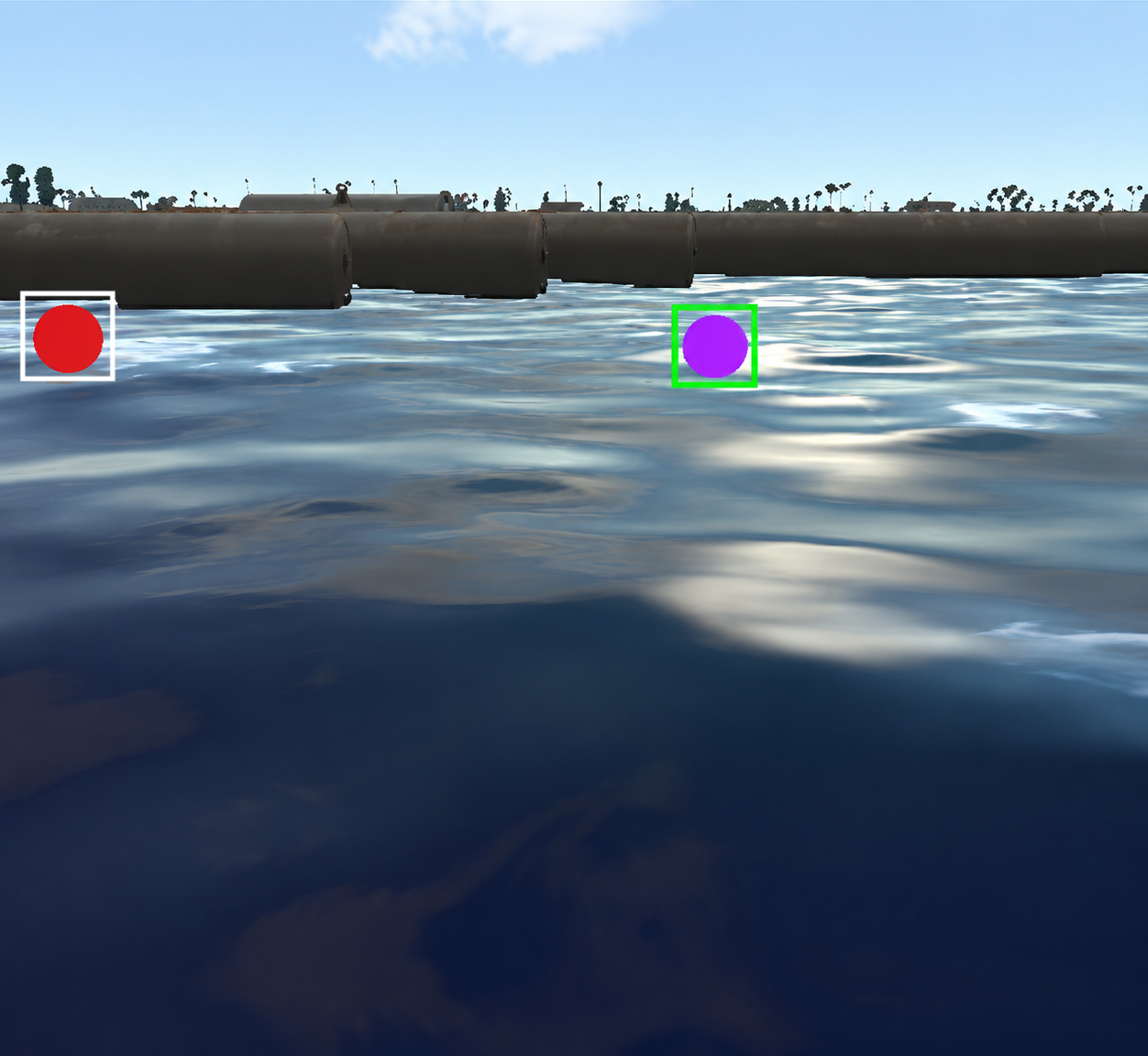}\\
\scriptsize Red ball: same ball shape but different colour.
\end{minipage}
&
\begin{minipage}{0.28\textwidth}
\centering
\includegraphics[width=\linewidth]{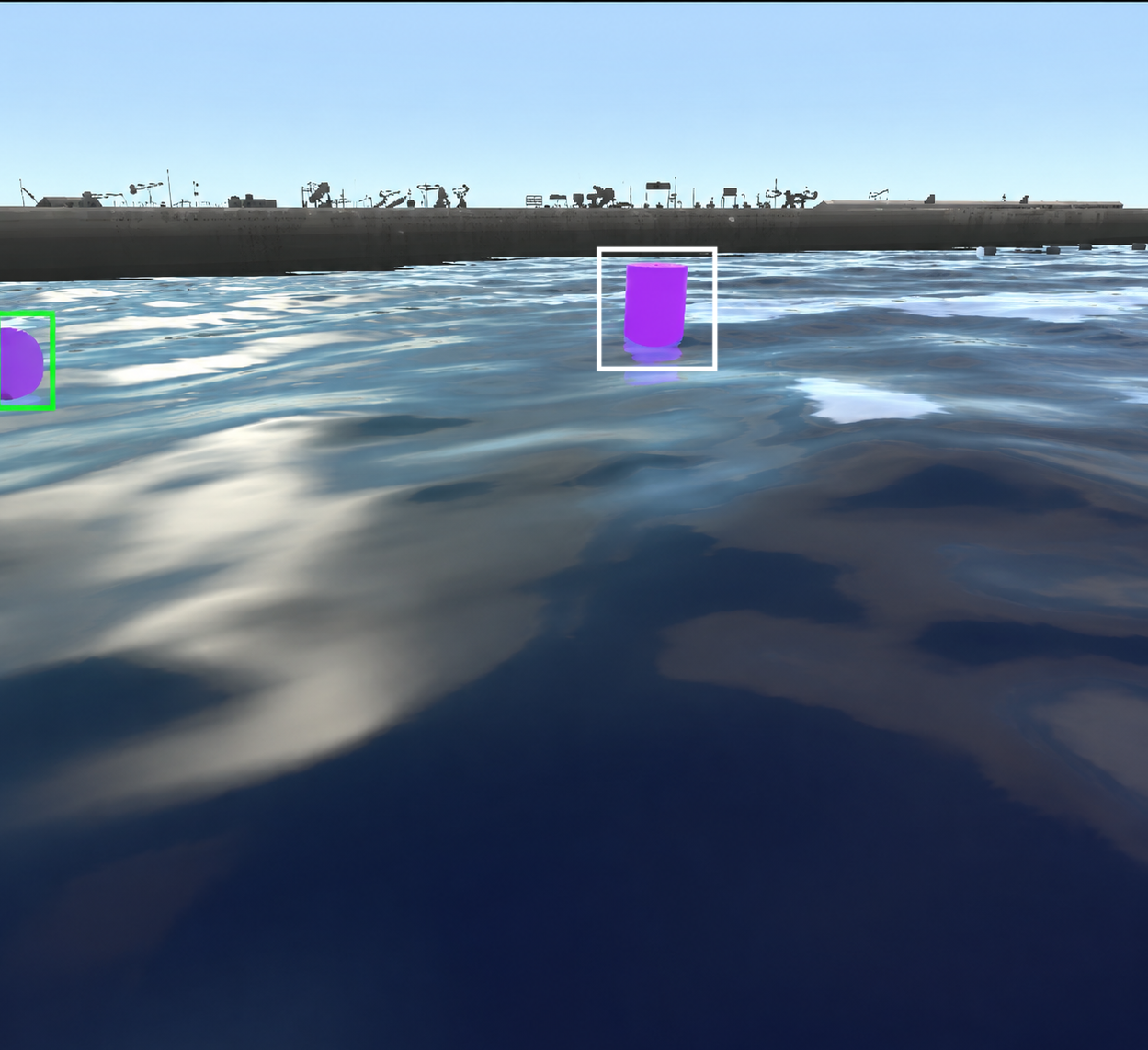}\\
\scriptsize Magenta cylinder: similar colour but different shape.
\end{minipage}
&
\begin{minipage}{0.28\textwidth}
\centering
\includegraphics[width=\linewidth]{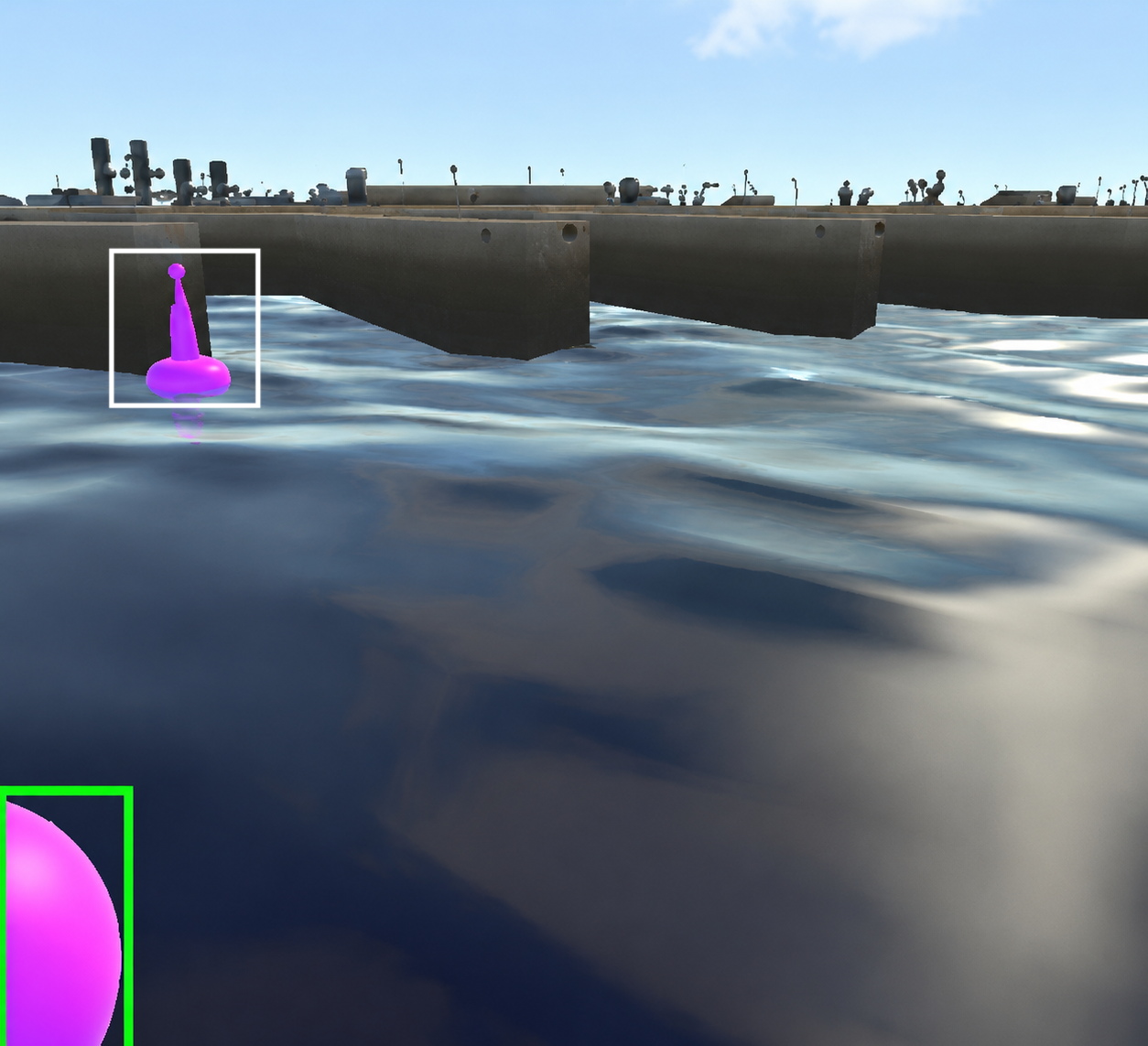}\\
\scriptsize Magenta buoy: similar object but not the target marker.
\end{minipage}
\\

\addlinespace[0.5em]

\textbf{Task~2} &
\begin{minipage}{0.28\textwidth}
\centering
\includegraphics[width=\linewidth]{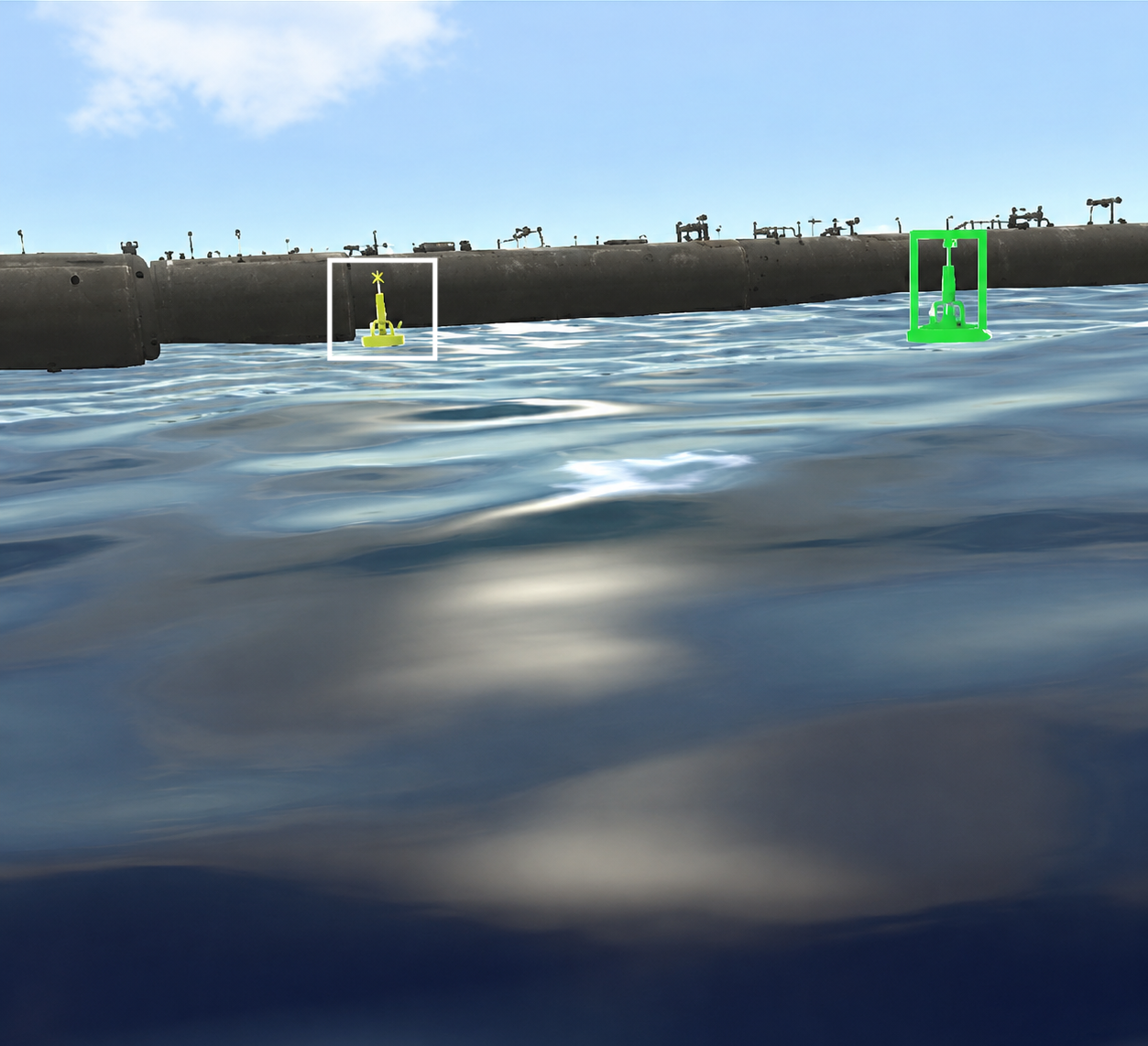}\\
\scriptsize Large yellow buoy: similar buoy structure but different colour.
\end{minipage}
&
\begin{minipage}{0.28\textwidth}
\centering
\includegraphics[width=\linewidth]{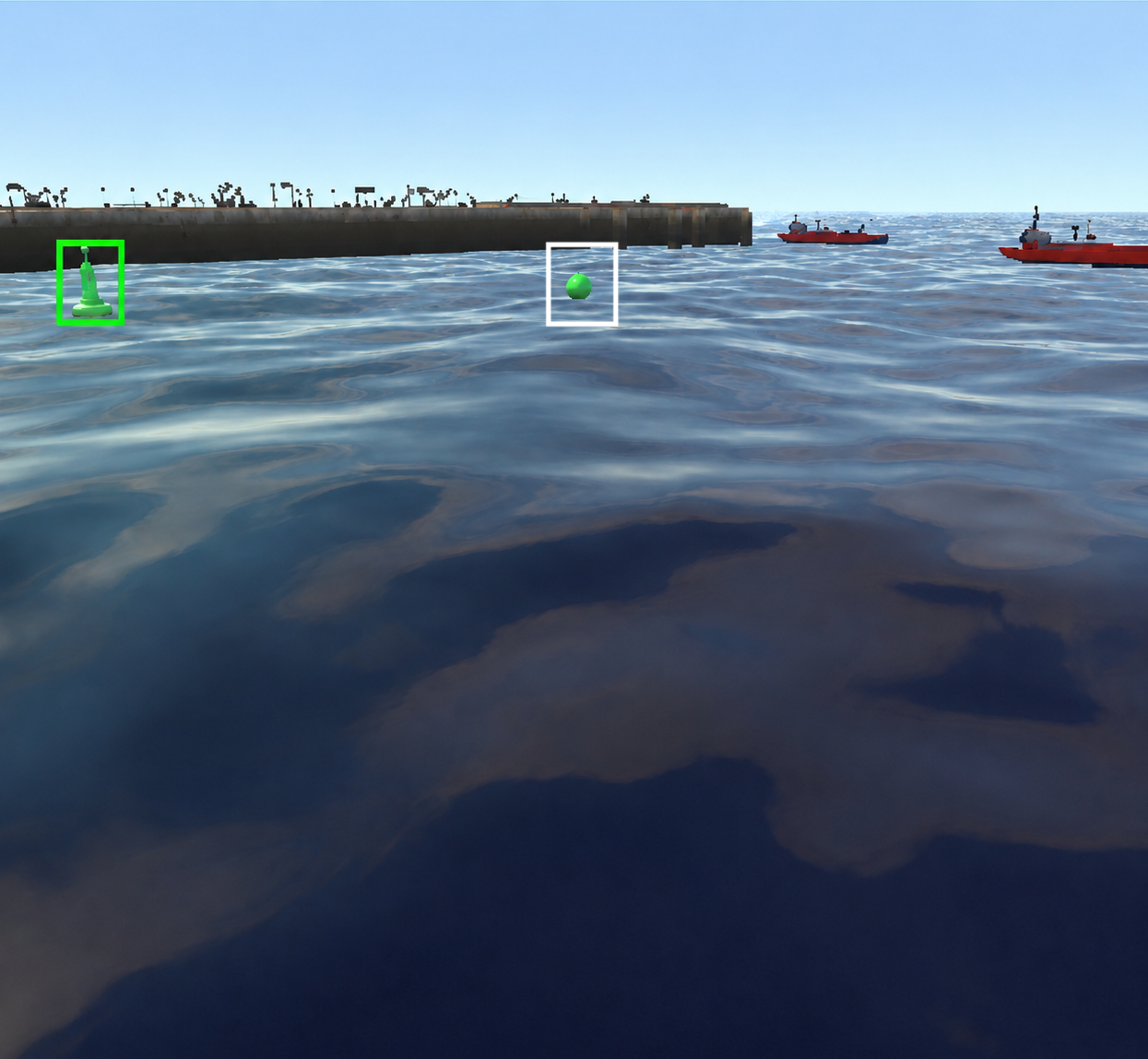}\\
\scriptsize Large green ball: similar colour and size, but not a buoy.
\end{minipage}
&
\begin{minipage}{0.28\textwidth}
\centering
\includegraphics[width=\linewidth]{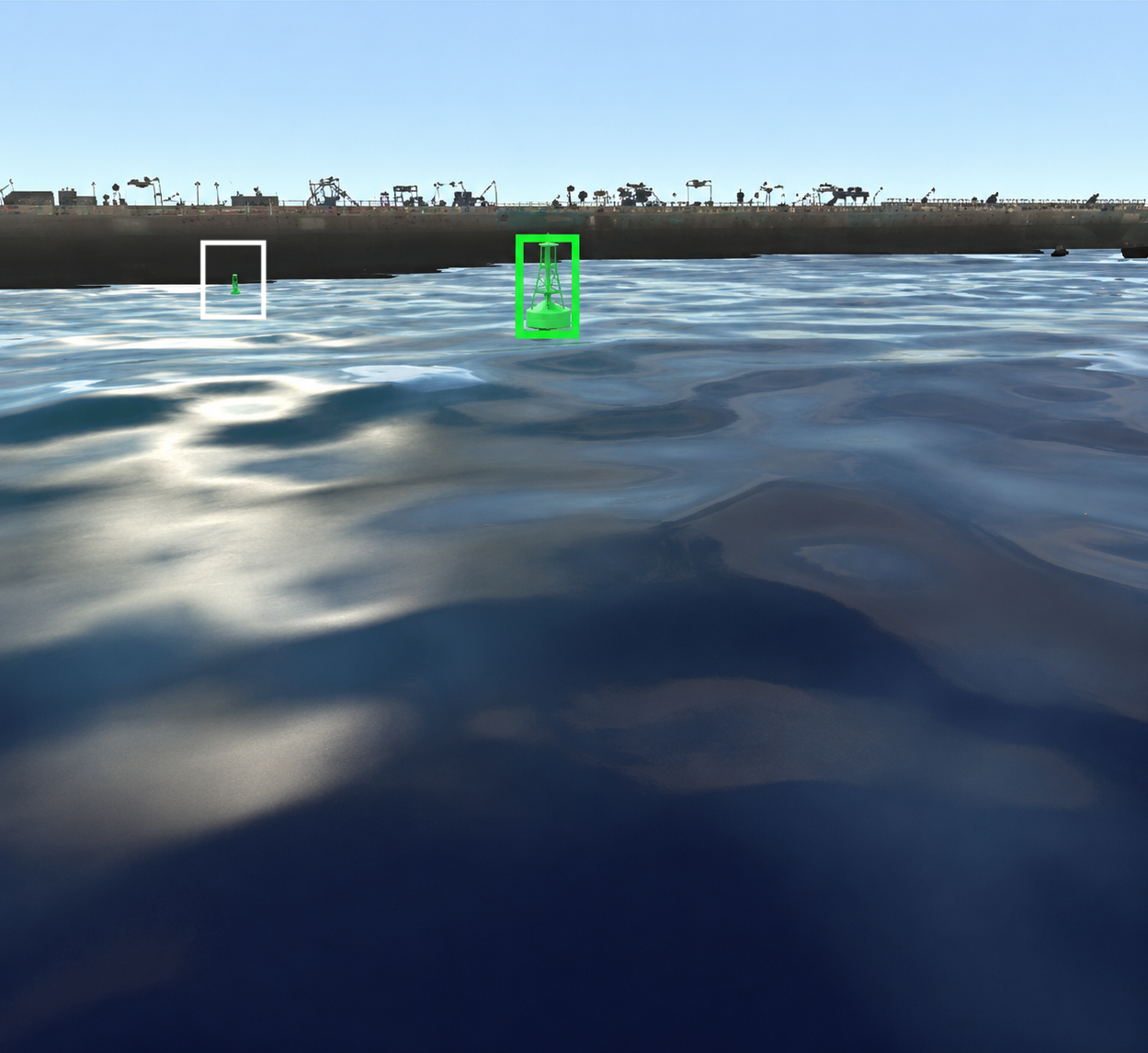}\\
\scriptsize Small green buoy: green buoy but not the specified large buoy.
\end{minipage}
\\

\addlinespace[0.5em]

\textbf{Task~3} &
\begin{minipage}{0.28\textwidth}
\centering
\includegraphics[width=\linewidth]{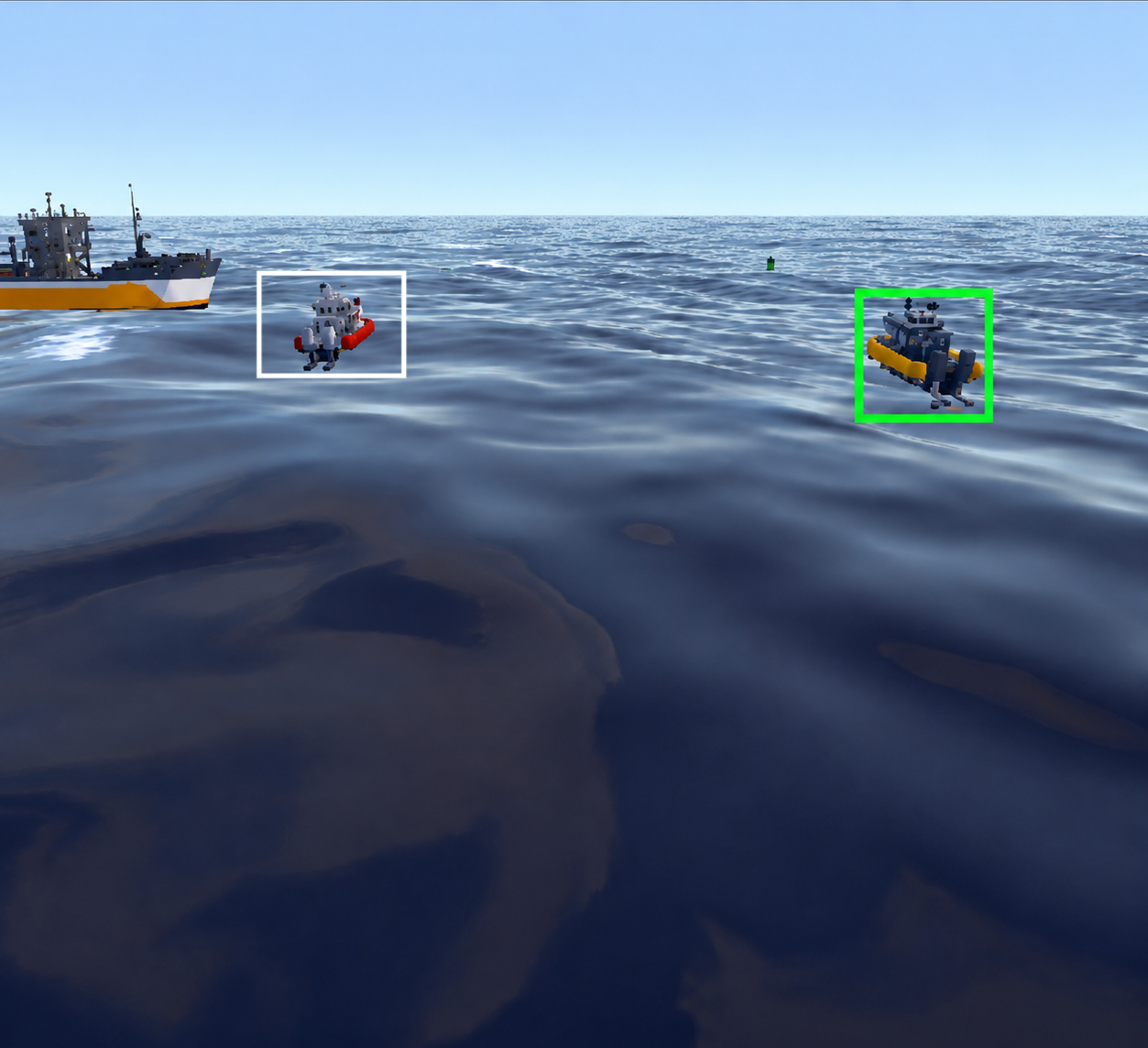}\\
\scriptsize Red patrol boat: same vessel type but different colour.
\end{minipage}
&
\begin{minipage}{0.28\textwidth}
\centering
\includegraphics[width=\linewidth]{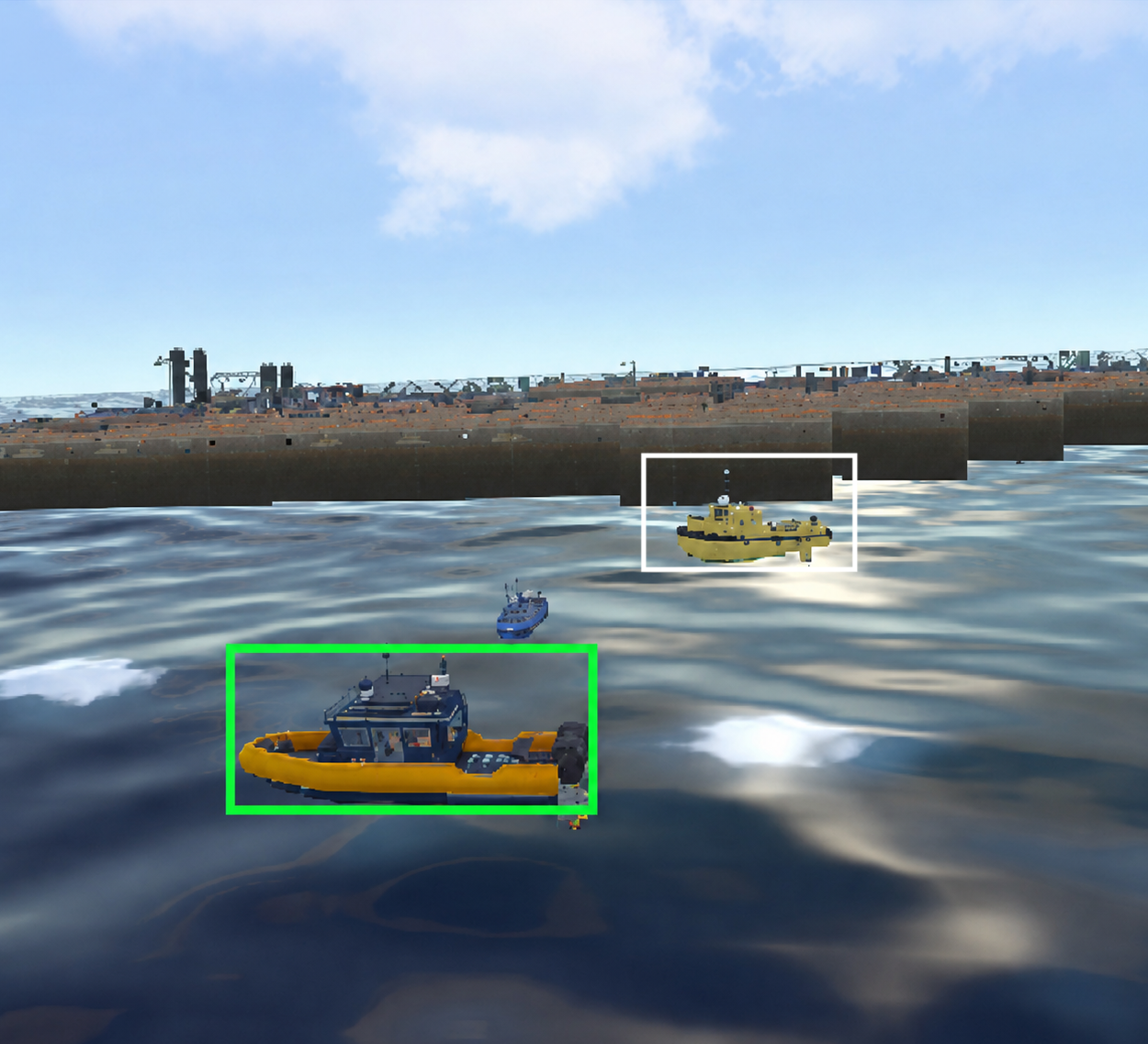}\\
\scriptsize Yellow tugboat: similar yellow vessel but different category.
\end{minipage}
&
\begin{minipage}{0.28\textwidth}
\centering
\includegraphics[width=\linewidth]{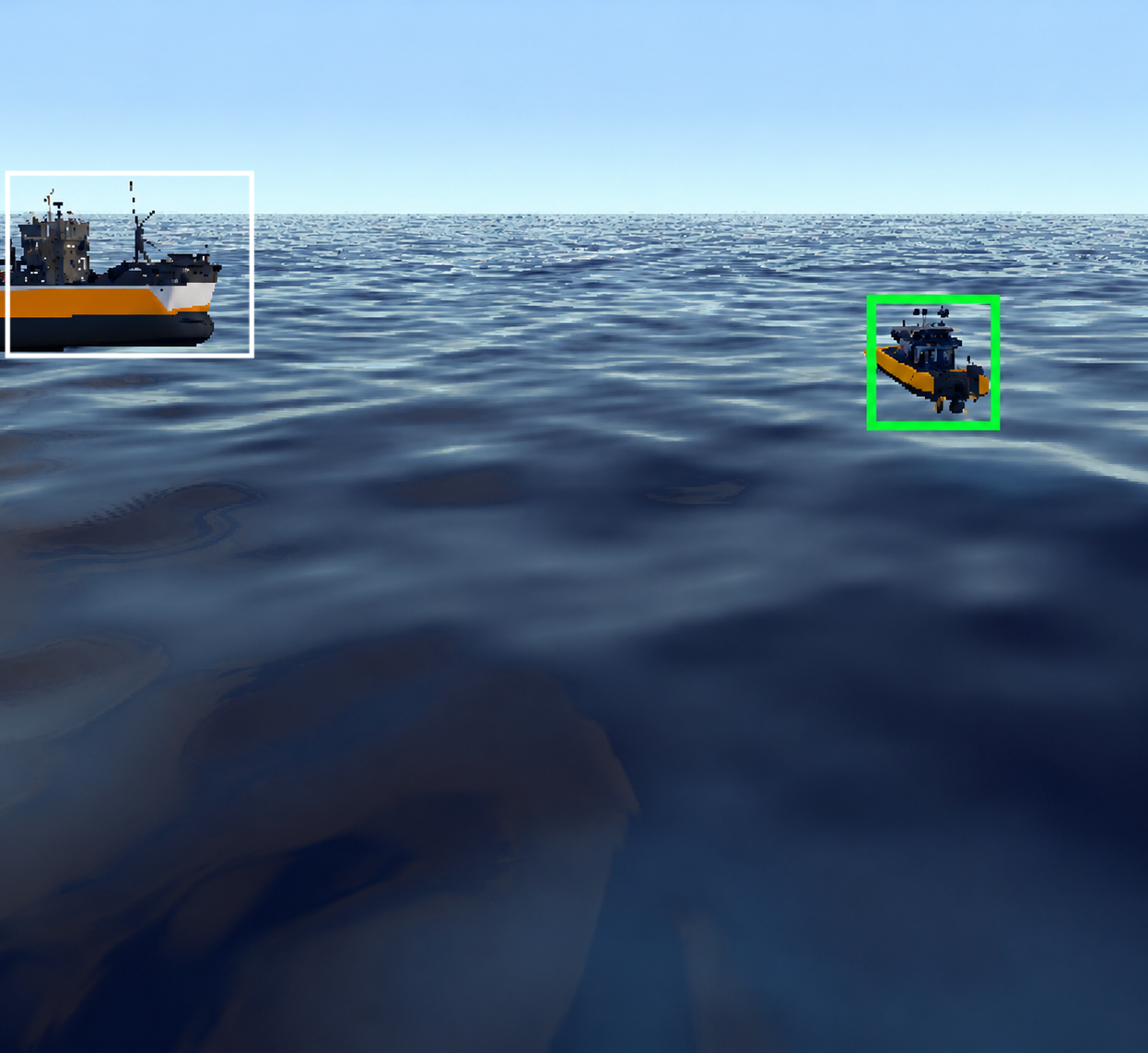}\\
\scriptsize Yellow container ship: maritime vessel but not a patrol boat.
\end{minipage}
\\

\bottomrule
\end{tabular}

\caption{
Qualitative examples of controlled distractor settings in Experiment~3. Rows correspond to Tasks~1, 2, and~3, while columns correspond to colour, shape, and semantic distractors, respectively. Green boxes indicate instructed targets, and white boxes indicate distractors.
}
\label{fig:exp3_distractor_grid}
\end{figure*}

Figures~\ref{fig:exp3_distractor_grid} show representative successful cases under colour, shape, and semantic distractors. The results visually confirm that SGNav can still select the intended target rather than the distractor, although semantic distractors remain more challenging because they share more direct task-level meaning with the target.

\subsection{Results of Experiment 4}

Table~\ref{tab:exp4_layout} reports the cross-layout generalisation results for Task~1, where all methods are trained on Layout~A and evaluated on held-out Layouts~B--D. SGNav achieves high SR across the held-out layouts, with $98.7\pm1.5\%$, $98.3\pm1.5\%$, and $97.7\pm2.1\%$ in Layouts~B, C, and D, respectively. The STA values remain around $67$--$70\%$, suggesting that frame-level grounding becomes more difficult under layout-induced changes in viewpoint, target appearance, and surrounding distractors. Nevertheless, the low ColR and high SR indicate that SGNav can still use the grounded target cues to support closed-loop navigation across held-out port layouts.

In contrast, Vision-Only PPO and No-Target PPO fail to achieve meaningful success in the held-out layouts, showing that navigation without explicit semantic target grounding cannot reliably handle spatial configuration changes. Oracle PPO achieves the highest transfer performance because it receives privileged target information, and therefore serves as an upper-performance bound rather than a directly comparable deployable method. Compared with Oracle PPO, SGNav requires longer CT and obtains lower PE, since it must ground the target online from onboard visual observations and adjust its trajectory under the changed layout. As the spatial variation becomes more challenging from Layout~B to Layout~D, SGNav generally requires longer completion time and produces lower path efficiency, indicating that more complex port geometries impose additional trajectory-correction demands.

\begin{table*}[t]
\centering
\caption{Cross-layout generalisation results for Task~1. All methods are trained on Layout~A and evaluated on held-out layouts. Results are reported as mean $\pm$ standard deviation over three random seeds.}
\label{tab:exp4_layout}
\renewcommand{\arraystretch}{1.12}
\setlength{\tabcolsep}{4pt}
\small
\begin{tabular}{llccccc}
\toprule
Method & Eval. layout  
& SR (\%) 
& STA (\%) 
& ColR (\%) 
& CT (sim. s) 
& PE \\
\midrule

Vision-Only PPO 
& Layout~B 
& $0.3 \pm 0.6$ 
& N/A 
& $2.0 \pm 1.0$ 
& N/A 
& N/A \\
& Layout~C 
& $0.0 \pm 0.0$ 
& N/A 
& $0.3 \pm 0.6$ 
& N/A 
& N/A \\
& Layout~D 
& $0.0 \pm 0.0$ 
& N/A 
& $0.3 \pm 0.6$ 
& N/A 
& N/A \\

\midrule

No-Target PPO 
& Layout~B  
& $0.0 \pm 0.0$ 
& N/A 
& $0.3 \pm 0.6$ 
& N/A 
& N/A \\
& Layout~C  
& $0.0 \pm 0.0$ 
& N/A 
& $0.0 \pm 0.0$ 
& N/A 
& N/A \\
& Layout~D  
& $0.0 \pm 0.0$ 
& N/A 
& $0.3 \pm 0.6$ 
& N/A 
& N/A \\

\midrule

Oracle PPO 
& Layout~B  
& $99.7 \pm 0.6$ 
& N/A 
& $0.0 \pm 0.0$ 
& $468.2 \pm 18.7$ 
& $1.003 \pm 0.021$ \\
& Layout~C  
& $99.3 \pm 1.2$ 
& N/A 
& $0.0 \pm 0.0$ 
& $501.7 \pm 21.4$ 
& $1.001 \pm 0.018$ \\
& Layout~D  
& $99.0 \pm 1.0$ 
& N/A 
& $0.3 \pm 0.6$ 
& $584.1 \pm 26.9$ 
& $1.000 \pm 0.020$ \\

\midrule

\textbf{Proposed SGNav} 
& Layout~B 
& $\mathbf{98.7 \pm 1.5}$ 
& $\mathbf{70.5 \pm 2.4}$ 
& $\mathbf{0.3 \pm 0.6}$ 
& $\mathbf{888.2 \pm 34.6}$ 
& $\mathbf{0.981 \pm 0.024}$ \\
& Layout~C 
& $\mathbf{98.3 \pm 1.5}$ 
& $\mathbf{67.4 \pm 2.8}$ 
& $\mathbf{0.3 \pm 0.6}$ 
& $\mathbf{969.4 \pm 41.2}$  
& $\mathbf{0.949 \pm 0.027}$ \\
& Layout~D 
& $\mathbf{97.7 \pm 2.1}$ 
& $\mathbf{69.5 \pm 2.6}$ 
& $\mathbf{0.7 \pm 0.6}$ 
& $\mathbf{1090.2 \pm 48.5}$ 
& $\mathbf{0.897 \pm 0.031}$ \\

\bottomrule
\end{tabular}

\vspace{0.5em}
\begin{minipage}{0.96\textwidth}
\footnotesize
\textit{Note:} N/A indicates that the metric is not applicable. STA is not reported for Vision-Only PPO, No-Target PPO, and Oracle PPO because these baselines do not perform explicit language-grounded target resolution. CT and PE are reported only for successful episodes and are therefore N/A when no successful episode is achieved. CT denotes simulated episode completion time rather than wall-clock training or inference time.
\end{minipage}
\end{table*}

\begin{figure}
    \centering
    \includegraphics[width=\linewidth]{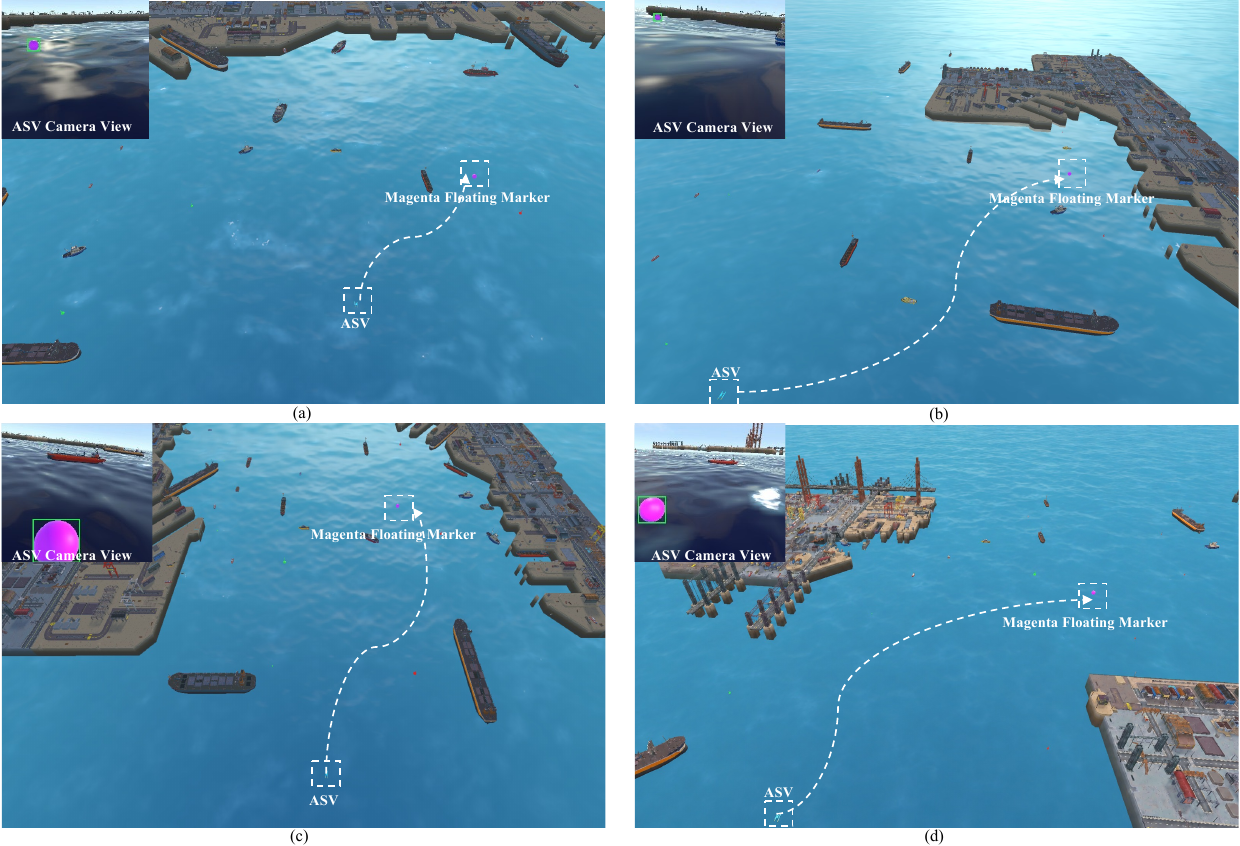}
    \caption{
    Representative trajectories and onboard observatios in the training and held-out layouts. Panels (a)--(d) show SGNav trajectories and onboard camera grounding views in Layouts~A, B, C, and D, respectively. Layout~A is the training layout, while Layouts~B, C, and D are held-out evaluation layouts. In each layout, the ASV follows a target-directed trajectory towards the instructed Magenta Floating Marker under different port configurations.
    }
    \label{fig:exp4_alllayouts_traj}
\end{figure}

Figure~\ref{fig:exp4_alllayouts_traj} further visualises the layout-transfer results. Layout~A shows the training layout, while Layouts~B--D show held-out evaluation layouts with different spatial configurations. Across these layouts, SGNav produces target-directed trajectories towards the Magenta Floating Marker despite changes in berth arrangement, vessel distribution, obstacle placement, and target location. The onboard camera views further illustrate that the instructed target can be visually grounded during navigation. These qualitative results suggest that SGNav does not simply memorise the training trajectory, but uses language-grounded target perception to support navigation across spatially varied port layouts.

\subsection{Results of Experiment 5}

Table~\ref{tab:ablation_results} reports the ablation results on Tasks~1 and~3. The Full SGNav results are aligned with the main task evaluation in Table~\ref{tab:exp1_results}, where the No-Target PPO and Vision-Only PPO baselines have already shown the necessity of target grounding. Therefore, this ablation focuses on two non-trivial components of the language-grounded target resolution module: harbour-aware filtering and semantic consistency.

The effects of these two modules differ across the two tasks. In Task~1, removing harbour-aware filtering reduces SR from $97.0\pm1.2\%$ to $93.6\pm1.7\%$ and increases WTR from $1.41\pm0.64\%$ to $6.4\pm1.5\%$. Removing semantic consistency has only a minor effect on SR, while STA and DRR remain high. This suggests that the controlled Magenta Floating Marker task mainly depends on successful target grounding, with limited ambiguity after the target is detected.

In Task~3, the impact of these modules is much stronger. Removing harbour-aware filtering reduces SR from $90.0\pm1.8\%$ to $40.4\pm2.6\%$, while removing semantic consistency reduces SR to $50.4\pm3.1\%$. The corresponding increases in WTR and decreases in DRR indicate that the ASV is more likely to select or follow misleading vessel-related targets when filtering or consistency checking is removed. These results show that visually complex maritime targets require not only grounding, but also distractor suppression and semantic consistency mechanisms for reliable target discrimination.

\begin{table*}[t]
\centering
\caption{Ablation study results on Tasks~1 and~3. Results are reported as mean $\pm$ standard deviation over three random seeds.}
\label{tab:ablation_results}
\renewcommand{\arraystretch}{1.12}
\setlength{\tabcolsep}{4pt}
\small
\begin{tabular}{llcccc}
\toprule
Task & Variant 
& SR (\%) 
& STA (\%) 
& WTR (\%) 
& DRR (\%) \\
\midrule

\multirow{3}{*}{Task~1}
& \textbf{Full SGNav}          
& $\mathbf{97.0 \pm 1.2}$ 
& $\mathbf{98.59 \pm 0.64}$ 
& $\mathbf{1.41 \pm 0.64}$ 
& $\mathbf{98.6 \pm 0.8}$ \\

& w/o Harbour-Aware Filtering       
& $93.6 \pm 1.7$ 
& $93.6 \pm 1.5$ 
& $6.4 \pm 1.5$ 
& $93.8 \pm 1.8$ \\

& w/o Semantic Consistency     
& $95.6 \pm 1.4$ 
& $99.6 \pm 0.4$ 
& $0.4 \pm 0.4$ 
& $99.2 \pm 0.7$ \\

\midrule

\multirow{3}{*}{Task~3}
& \textbf{Full SGNav}          
& $\mathbf{90.0 \pm 1.8}$
& $\mathbf{97.65 \pm 0.72}$
& $\mathbf{2.35 \pm 0.72}$
& $\mathbf{97.6 \pm 0.9}$ \\

& w/o Harbour-Aware Filtering       
& $40.4 \pm 2.6$ 
& $85.2 \pm 2.4$ 
& $14.8 \pm 2.4$ 
& $85.0 \pm 2.7$ \\

& w/o Semantic Consistency     
& $50.4 \pm 3.1$ 
& $52.4 \pm 3.0$ 
& $47.6 \pm 3.0$ 
& $52.8 \pm 3.2$ \\

\bottomrule
\end{tabular}

\vspace{0.5em}
\begin{minipage}{0.96\textwidth}
\footnotesize
\textit{Note:} Full SGNav follows the same evaluation protocol as Table~\ref{tab:exp1_results}. The necessity of target grounding is evaluated separately through the No-Target PPO and Vision-Only PPO baselines in Table~\ref{tab:exp1_results}.
\end{minipage}
\end{table*}


\subsection{Discussion}

The experimental results show that SGNav can support language-grounded semantic target navigation in visually complex maritime environments. Instead of relying on fixed coordinates or predefined target identifiers, SGNav connects an operator-provided target description with visual target grounding and downstream ASV control. The results across target-reaching tasks, instruction variants, distractor settings, and held-out port layouts suggest that semantic grounding provides an effective interface between language-level target specification and closed-loop maritime navigation.

A key finding is that language-grounded target resolution is essential for the proposed framework. SGNav achieves performance close to the Oracle PPO upper bound in the main task evaluation, while clearly outperforming Vision-Only PPO and No-Target PPO. The ablation study further confirms this role: removing the grounding module leads to complete navigation failure because the policy no longer receives a valid target-conditioned control state. This indicates that the downstream navigation policy alone is insufficient when the target is specified by language.

Harbour-aware filtering and semantic consistency become more important as the visual and semantic ambiguity increases. For the controlled Magenta Floating Marker task, successful grounding already provides a relatively clear target cue, and removing filtering or consistency only causes limited degradation. In contrast, the Yellow Patrol Boat task involves stronger interference from other vessel classes, viewpoint changes, and maritime background clutter. In this setting, filtering irrelevant candidates and maintaining semantic consistency improve target discrimination, reduce wrong-target failures, and stabilise navigation behaviour.

The ablation results also show that semantic navigation should not be evaluated using a single metric. A variant may retain relatively high semantic target accuracy on the subset of successful episodes while still producing a much lower success rate. Therefore, STA should be interpreted together with SR, WTR, and DRR. Target correctness among successful trials does not necessarily imply robust task completion, especially when the method frequently fails to reach the target or is distracted by competing objects.

The instruction generalisation and cross-layout results further indicate that SGNav is not limited to fixed command templates or a single port layout. The framework remains effective under synonym substitution, attribute-enriched descriptions, context-aware descriptions, longer natural-language instructions, and held-out spatial configurations. However, grounding accuracy is still affected by increased linguistic complexity, and SGNav requires longer completion time and lower path efficiency than Oracle PPO under layout changes. This suggests that language-grounded target specification improves flexibility and transferability, but more advanced language understanding and long-horizon planning remain important directions for future work.

The runtime analysis shows that the downstream PPO policy is lightweight, while the main computational cost comes from GroundingDINO candidate generation and CLIP-based semantic verification. With cached semantic updates, the current implementation can support closed-loop simulation control, but practical deployment would benefit from further acceleration through asynchronous perception-control execution, lower-frequency semantic grounding, model compression, and lightweight maritime grounding models.

Several limitations remain. First, the current implementation relies on RGB visual observations for target grounding. Although RGB cameras provide rich semantic information, camera-only perception can be unreliable in real port environments under poor illumination, rain, fog, glare, occlusion, or long-range observations. Future work should therefore integrate additional maritime sensing and information sources, such as radar, AIS, electronic nautical charts, LiDAR, thermal cameras, and vessel motion information, to improve robustness under adverse operational conditions. Second, all experiments are conducted in simulation. Real-world deployment will involve stronger sensor noise, dynamic vessel behaviour, environmental uncertainty, and imperfect calibration between perception and control. Sim-to-real validation, hardware-in-the-loop testing, and field experiments are therefore required before practical deployment. Third, the current semantic vocabulary and task set remain limited to object-level target-reaching tasks. Future work will investigate richer relational commands, temporal instructions, rule-aware navigation, and multi-agent semantic collaboration for more complex autonomous port operations.

Finally, this study focuses on whether language descriptions can be grounded to the correct maritime target and translated into closed-loop ASV navigation. Although language-based target descriptions provide a more flexible goal-specification mechanism than coordinates or fixed target identifiers, formal evaluation of operator workload, usability, and command preference is left for future work.

\FloatBarrier

\section{Conclusion}
\label{sec:6}
This study proposed Semantically Grounded Navigation (SGNav), a language-grounded target navigation framework for ASVs operating in harbour environments. The framework addresses the gap between language-based target specification and executable ASV control by integrating text-guided semantic grounding, harbour-aware candidate filtering, CLIP-based semantic verification, grounded target control-state construction, and downstream PPO-based policy execution. Instead of relying on predefined coordinates, reference trajectories, or fixed target identifiers, SGNav allows an ASV to associate an operator-provided target description with onboard visual observations and navigate towards the intended maritime target through closed-loop control.

Experiments in simulated port environments demonstrate that SGNav can support semantic target navigation across different target-reaching tasks, instruction variants, distractor settings, and held-out port layouts. Compared with Vision-Only PPO and No-Target PPO, SGNav achieves more reliable task completion and stronger target correctness, while the ablation study confirms the importance of language-grounded target resolution, harbour-aware filtering, and semantic consistency. The results also show that SGNav is not restricted to fixed command templates and can transfer to spatially varied port layouts by grounding the target from visual-semantic cues.

The findings suggest that language-grounded target specification can improve the flexibility and adaptability of ASV navigation, but several challenges remain. These include reduced grounding reliability under complex or ambiguous instructions, additional computational cost from online semantic perception, and the limitations of RGB-only perception in real port environments. Future work will focus on sim-to-real validation, onboard runtime optimisation, integration of maritime sensors such as radar, AIS, electronic charts, and LiDAR, and richer relational, temporal, rule-aware, and multi-agent semantic reasoning for complex autonomous port operations.

\newpage
\bibliographystyle{plainnat}
\bibliography{ref}

@inproceedings{anderson2018vision,
  title={Vision-and-language navigation: Interpreting visually-grounded navigation instructions in real environments},
  author={Anderson, Peter and Wu, Qi and Teney, Damien and Bruce, Jake and Johnson, Mark and S{\"u}nderhauf, Niko and Reid, Ian and Gould, Stephen and Van Den Hengel, Anton},
  booktitle={Proceedings of the IEEE conference on computer vision and pattern recognition},
  pages={3674--3683},
  year={2018}
}

@inproceedings{gu2022vision,
  title={Vision-and-language navigation: A survey of tasks, methods, and future directions},
  author={Gu, Jing and Stefani, Eliana and Wu, Qi and Thomason, Jesse and Wang, Xin Eric},
  booktitle={Proceedings of the 60th Annual Meeting of the Association for Computational Linguistics (Volume 1: Long Papers)},
  pages={7606--7623},
  year={2022}
}

@inproceedings{liu2024grounding,
  title={Grounding dino: Marrying dino with grounded pre-training for open-set object detection},
  author={Liu, Shilong and Zeng, Zhaoyang and Ren, Tianhe and Li, Feng and Zhang, Hao and Yang, Jie and Jiang, Qing and Li, Chunyuan and Yang, Jianwei and Su, Hang and others},
  booktitle={European conference on computer vision},
  pages={38--55},
  year={2024},
  organization={Springer}
}

@inproceedings{radford2021learning,
  title={Learning transferable visual models from natural language supervision},
  author={Radford, Alec and Kim, Jong Wook and Hallacy, Chris and Ramesh, Aditya and Goh, Gabriel and Agarwal, Sandhini and Sastry, Girish and Askell, Amanda and Mishkin, Pamela and Clark, Jack and others},
  booktitle={International conference on machine learning},
  pages={8748--8763},
  year={2021},
  organization={PmLR}
}

@article{schulman2017proximal,
  title={Proximal policy optimization algorithms},
  author={Schulman, John and Wolski, Filip and Dhariwal, Prafulla and Radford, Alec and Klimov, Oleg},
  journal={arXiv preprint arXiv:1707.06347},
  year={2017}
}

@article{lin2025multi,
  title={A multi-objective deep reinforcement learning framework for energy efficiency of autonomous harbor crafts},
  author={Lin, Yuqing and Xin, Jinghao and Zhang, Rangya and Yuen, Kum Fai},
  journal={Applied Energy},
  volume={401},
  pages={126809},
  year={2025},
  publisher={Elsevier}
}

@article{lin2025multiple,
  title={Multiple unmanned surface vehicles pathfinding in dynamic environment},
  author={Lin, Yuqing and Du, Liang and Yuen, Kum Fai},
  journal={Applied Soft Computing},
  volume={172},
  pages={112820},
  year={2025},
  publisher={Elsevier}
}

@article{lin2025machine,
  title={Machine learning applications for risk assessment in maritime transport: Current status and future directions},
  author={Lin, Yuqing and Li, Xue and Yuen, Kum Fai},
  journal={Engineering Applications of Artificial Intelligence},
  volume={155},
  pages={110959},
  year={2025},
  publisher={Elsevier}
}

@article{yao2026improving,
  title={Improving localization precision in open-vocabulary object detection through reinforcement learning-based model collaboration},
  author={Yao, Xudong and Jiang, Han and Liu, Hao and Yang, Xiaoshan},
  journal={Multimedia Systems},
  volume={32},
  number={4},
  pages={249},
  year={2026},
  publisher={Springer}
}

@inproceedings{yu2023fusing,
  title={Fusing pre-trained language models with multimodal prompts through reinforcement learning},
  author={Yu, Youngjae and Chung, Jiwan and Yun, Heeseung and Hessel, Jack and Park, Jae Sung and Lu, Ximing and Zellers, Rowan and Ammanabrolu, Prithviraj and Le Bras, Ronan and Kim, Gunhee and others},
  booktitle={Proceedings of the IEEE/CVF Conference on Computer Vision and Pattern Recognition},
  pages={10845--10856},
  year={2023}
}

@article{son2024teacher,
  title={Teacher--student model using grounding DINO and you only look once for multi-sensor-based object detection},
  author={Son, Jinhwan and Jung, Heechul},
  journal={Applied Sciences},
  volume={14},
  number={6},
  pages={2232},
  year={2024},
  publisher={MDPI}
}

@inproceedings{wu2023cora,
  title={Cora: Adapting clip for open-vocabulary detection with region prompting and anchor pre-matching},
  author={Wu, Xiaoshi and Zhu, Feng and Zhao, Rui and Li, Hongsheng},
  booktitle={Proceedings of the IEEE/CVF conference on computer vision and pattern recognition},
  pages={7031--7040},
  year={2023}
}

@inproceedings{jiang2024visual,
  title={Visual Grounding for Object-Level Generalization in Reinforcement Learning},
  author={Jiang, Haobin and Lu, Zongqing},
  booktitle={European Conference on Computer Vision},
  year={2024}
}

@article{luo2024learning,
  title={Learning multimodal adaptive relation graph and action boost memory for visual navigation},
  author={Luo, Jian and Cai, Bo and Yu, Yaoxiang and Ke, Aihua and Zhou, Kang and Zhang, Jian},
  journal={Advanced Engineering Informatics},
  volume={62},
  pages={102678},
  year={2024},
  publisher={Elsevier}
}

@article{kim2022bim,
  title={BIM-based semantic building world modeling for robot task planning and execution in built environments},
  author={Kim, Kyungki and Peavy, Matthew},
  journal={Automation in Construction},
  volume={138},
  pages={104247},
  year={2022},
  publisher={Elsevier}
}

@article{yang2024hogn,
  title={HOGN-TVGN: Human-inspired embodied object goal navigation based on time-varying knowledge graph inference networks for robots},
  author={Yang, Baojiang and Yuan, Xianfeng and Ying, Zhongmou and Zhang, Jialin and Song, Boyi and Song, Yong and Zhou, Fengyu and Sheng, Weihua},
  journal={Advanced Engineering Informatics},
  volume={62},
  pages={102671},
  year={2024},
  publisher={Elsevier}
}

@article{yang2024enhanced,
  title={Enhanced visual SLAM for construction robots by efficient integration of dynamic object segmentation and scene semantics},
  author={Yang, Liu and Cai, Hubo},
  journal={Advanced Engineering Informatics},
  volume={59},
  pages={102313},
  year={2024},
  publisher={Elsevier}
}

@article{chen2022pathfinding,
  title={Pathfinding method for an indoor drone based on a BIM-semantic model},
  author={Chen, Qingxiang and Chen, Jing and Huang, Wumeng},
  journal={Advanced Engineering Informatics},
  volume={53},
  pages={101686},
  year={2022},
  publisher={Elsevier}
}

@article{paden2016survey,
  title={A survey of motion planning and control techniques for self-driving urban vehicles},
  author={Paden, Brian and {\v{C}}{\'a}p, Michal and Yong, Sze Zheng and Yershov, Dmitry and Frazzoli, Emilio},
  journal={IEEE Transactions on intelligent vehicles},
  volume={1},
  number={1},
  pages={33--55},
  year={2016},
  publisher={IEEE}
}

@inproceedings{saravanakumar2011waypoint,
  title={Waypoint Guidance based Planar Path Following and Obstacle Avoidance of Autonomous Underwater Vehicle.},
  author={Saravanakumar, S and Asokan, T},
  booktitle={ICINCO (2)},
  pages={191--198},
  year={2011}
}

@inproceedings{ku2020room,
  title={Room-across-room: Multilingual vision-and-language navigation with dense spatiotemporal grounding},
  author={Ku, Alexander and Anderson, Peter and Patel, Roma and Ie, Eugene and Baldridge, Jason},
  booktitle={Proceedings of the 2020 Conference on Empirical Methods in Natural Language Processing (EMNLP)},
  pages={4392--4412},
  year={2020}
}

@article{fried2018speaker,
  title={Speaker-follower models for vision-and-language navigation},
  author={Fried, Daniel and Hu, Ronghang and Cirik, Volkan and Rohrbach, Anna and Andreas, Jacob and Morency, Louis-Philippe and Berg-Kirkpatrick, Taylor and Saenko, Kate and Klein, Dan and Darrell, Trevor},
  journal={Advances in neural information processing systems},
  volume={31},
  year={2018}
}

@article{ma2019self,
  title={Self-monitoring navigation agent via auxiliary progress estimation},
  author={Ma, Chih-Yao and Lu, Jiasen and Wu, Zuxuan and AlRegib, Ghassan and Kira, Zsolt and Socher, Richard and Xiong, Caiming},
  journal={arXiv preprint arXiv:1901.03035},
  year={2019}
}

@inproceedings{wang2019reinforced,
  title={Reinforced cross-modal matching and self-supervised imitation learning for vision-language navigation},
  author={Wang, Xin and Huang, Qiuyuan and Celikyilmaz, Asli and Gao, Jianfeng and Shen, Dinghan and Wang, Yuan-Fang and Wang, William Yang and Zhang, Lei},
  booktitle={Proceedings of the IEEE/CVF conference on computer vision and pattern recognition},
  pages={6629--6638},
  year={2019}
}

@inproceedings{wang2021structured,
  title={Structured scene memory for vision-language navigation},
  author={Wang, Hanqing and Wang, Wenguan and Liang, Wei and Xiong, Caiming and Shen, Jianbing},
  booktitle={Proceedings of the IEEE/CVF conference on Computer Vision and Pattern Recognition},
  pages={8455--8464},
  year={2021}
}

@inproceedings{shah2023lmnav,
  title={Lm-nav: Robotic navigation with large pre-trained models of language, vision, and action},
  author={Shah, Dhruv and Osi{\'n}ski, B{\l}a{\.z}ej and Levine, Sergey and others},
  booktitle={Conference on robot learning},
  pages={492--504},
  year={2023},
  organization={pmlr}
}

@inproceedings{shah2023gnm,
  title={Gnm: A general navigation model to drive any robot},
  author={Shah, Dhruv and Sridhar, Ajay and Bhorkar, Arjun and Hirose, Noriaki and Levine, Sergey},
  booktitle={2023 IEEE International Conference on Robotics and Automation (ICRA)},
  pages={7226--7233},
  year={2023},
  organization={IEEE}
}

@inproceedings{shah2023vint,
  title={ViNT: A Foundation Model for Visual Navigation},
  author={Shah, Dhruv and Sridhar, Ajay and Dashora, Nitish and Stachowicz, Kyle and Black, Kevin and Hirose, Noriaki and Levine, Sergey},
  booktitle={Proceedings of The 7th Conference on Robot Learning},
  pages={711--733},
  year={2023},
  series={Proceedings of Machine Learning Research},
  volume={229},
  publisher={PMLR}
}

@inproceedings{sridhar2024nomad,
  title={Nomad: Goal masked diffusion policies for navigation and exploration},
  author={Sridhar, Ajay and Shah, Dhruv and Glossop, Catherine and Levine, Sergey},
  booktitle={2024 IEEE International Conference on Robotics and Automation (ICRA)},
  pages={63--70},
  year={2024},
  organization={IEEE}
}

@inproceedings{hirose2025lelan,
  title={LeLaN: Learning A Language-Conditioned Navigation Policy from In-the-Wild Video},
  author={Hirose, Noriaki and Glossop, Catherine and Sridhar, Ajay and Mees, Oier and Levine, Sergey},
  booktitle={Proceedings of The 8th Conference on Robot Learning},
  pages={666--688},
  year={2025},
  series={Proceedings of Machine Learning Research},
  volume={270},
  publisher={PMLR}
}

@inproceedings{khanna2024goat,
  title={Goat-bench: A benchmark for multi-modal lifelong navigation},
  author={Khanna, Mukul and Ramrakhya, Ram and Chhablani, Gunjan and Yenamandra, Sriram and Gervet, Theophile and Chang, Matthew and Kira, Zsolt and Chaplot, Devendra Singh and Batra, Dhruv and Mottaghi, Roozbeh},
  booktitle={Proceedings of the IEEE/CVF Conference on Computer Vision and Pattern Recognition},
  pages={16373--16383},
  year={2024}
}

@inproceedings{zhou2024navgpt,
  title={Navgpt: Explicit reasoning in vision-and-language navigation with large language models},
  author={Zhou, Gengze and Hong, Yicong and Wu, Qi},
  booktitle={Proceedings of the AAAI Conference on Artificial Intelligence},
  volume={38},
  number={7},
  pages={7641--7649},
  year={2024}
}

@inproceedings{zhou2024navgpt2,
  title={Navgpt-2: Unleashing navigational reasoning capability for large vision-language models},
  author={Zhou, Gengze and Hong, Yicong and Wang, Zun and Wang, Xin Eric and Wu, Qi},
  booktitle={European Conference on Computer Vision},
  pages={260--278},
  year={2024},
  organization={Springer}
}

@inproceedings{rana2023sayplan,
  title={SayPlan: Grounding Large Language Models using 3D Scene Graphs for Scalable Robot Task Planning},
  author={Rana, Krishan and Haviland, Jesse and Garg, Sourav and Jad Abou-Chakra, Ian Reid and Suenderhauf, Niko},
  booktitle={Proceedings of The 7th Conference on Robot Learning},
  pages={23--72},
  year={2023},
  series={Proceedings of Machine Learning Research},
  volume={229},
  publisher={PMLR}
}

@inproceedings{elnoor2025vlm,
  title={VLM-GroNav: Robot Navigation Using Physically Grounded Vision-Language Models in Outdoor Environments},
  author={Elnoor, Mohamed and Weerakoon, Kasun and Seneviratne, Gershom and Xian, Ruiqi and Guan, Tianrui and Jaffar, Mohamed Khalid M and Rajagopal, Vignesh and Manocha, Dinesh},
  booktitle={2025 IEEE International Conference on Robotics and Automation (ICRA)},
  pages={2391--2398},
  year={2025},
  organization={IEEE}
}

@inproceedings{wang2026expand,
  title={Expand Your Scope: Semantic Cognition over Potential-Based Exploration for Embodied Visual Navigation},
  author={Wang, Neng and Chen, Wei and Chen, Lin and Ji, Hong and Guo, Zhi and Zhang, Xue and Sun, Hao},
  booktitle={Proceedings of the AAAI Conference on Artificial Intelligence},
  volume={40},
  number={22},
  pages={18620--18628},
  year={2026}
}

@inproceedings{chaplot2020object,
  title={Object Goal Navigation using Goal-Oriented Semantic Exploration},
  author={Chaplot, Devendra Singh and Gandhi, Dhiraj Prakashchand and Gupta, Abhinav and Salakhutdinov, Ruslan},
  booktitle={Advances in Neural Information Processing Systems},
  volume={33},
  pages={4247--4258},
  year={2020}
}

@inproceedings{du2020learning,
  title={Learning Object Relation Graph and Tentative Policy for Visual Navigation},
  author={Du, Heming and Yu, Xin and Zheng, Liang},
  booktitle={Computer Vision -- ECCV 2020},
  pages={19--34},
  year={2020},
  publisher={Springer},
  doi={10.1007/978-3-030-58571-6_2}
}

@article{wang2024goal,
  title={Goal-Oriented Visual Semantic Navigation Using Semantic Knowledge Graph and Transformer},
  author={Wang, Z. and Tian, G.},
  journal={IEEE Transactions on Automation Science and Engineering},
  volume={22},
  pages={1647--1657},
  year={2024},
  publisher={IEEE}
}

@article{jiang2023learning,
  title={Learning Relation in Crowd Using Gated Graph Convolutional Networks for DRL-Based Robot Navigation},
  author={Jiang, H. and Bhujel, N. and Lin, Z. and Wan, K. W. and Li, J. and Jayavelu, S. and Jiang, X.},
  journal={IEEE Transactions on Intelligent Transportation Systems},
  volume={25},
  number={6},
  pages={5085--5095},
  year={2023},
  publisher={IEEE}
}

@inproceedings{kiran2022spatial,
  title={Spatial Relation Graph and Graph Convolutional Network for Object Goal Navigation},
  author={Kiran, D. S. and Anand, K. and Kharyal, C. and Kumar, G. and Gireesh, N. and Banerjee, S. and Krishna, M.},
  booktitle={Proceedings of the 2022 IEEE 18th International Conference on Automation Science and Engineering},
  pages={1392--1398},
  year={2022},
  organization={IEEE}
}

@inproceedings{wang2023gridmm,
  title={GridMM: Grid Memory Map for Vision-and-Language Navigation},
  author={Wang, Zihan and Li, Xiang and Yang, Jiahao and Liu, Yicheng and Jiang, Shuqiang},
  booktitle={Proceedings of the IEEE/CVF International Conference on Computer Vision},
  pages={15625--15636},
  year={2023}
}

@inproceedings{kamath2021mdetr,
  title={MDETR: Modulated Detection for End-to-End Multi-Modal Understanding},
  author={Kamath, Aishwarya and Singh, Mannat and LeCun, Yann and Synnaeve, Gabriel and Misra, Ishan and Carion, Nicolas},
  booktitle={Proceedings of the IEEE/CVF International Conference on Computer Vision},
  pages={1780--1790},
  year={2021}
}

@inproceedings{li2022grounded,
  title={Grounded Language-Image Pre-Training},
  author={Li, Liunian Harold and Zhang, Pengchuan and Zhang, Haotian and Yang, Jianwei and Li, Chunyuan and Zhong, Yiwu and Wang, Lijuan and Yuan, Lu and Zhang, Lei and Hwang, Jenq-Neng and Chang, Kai-Wei and Gao, Jianfeng},
  booktitle={Proceedings of the IEEE/CVF Conference on Computer Vision and Pattern Recognition},
  pages={10965--10975},
  year={2022}
}

@inproceedings{zhang2022glipv2,
  title={GLIPv2: Unifying Localization and Vision-Language Understanding},
  author={Zhang, Haotian and Zhang, Pengchuan and Hu, Xiaowei and Chen, Yen-Chun and Li, Liunian Harold and Dai, Xiyang and Wang, Lijuan and Yuan, Lu and Hwang, Jenq-Neng and Gao, Jianfeng},
  booktitle={Advances in Neural Information Processing Systems},
  volume={35},
  pages={36067--36080},
  year={2022}
}

@inproceedings{minderer2022simple,
  title={Simple Open-Vocabulary Object Detection with Vision Transformers},
  author={Minderer, Matthias and Gritsenko, Alexey and Stone, Austin and Neumann, Maxim and Weissenborn, Dirk and Dosovitskiy, Alexey and Mahendran, Aravindh and Arnab, Anurag and Dehghani, Mostafa and Shen, Zhuoran and Wang, Xiao and Zhai, Xiaohua and Kipf, Thomas and Houlsby, Neil},
  booktitle={Computer Vision -- ECCV 2022},
  pages={728--755},
  year={2022},
  publisher={Springer}
}

@inproceedings{kuo2023fvlm,
  title={F-VLM: Open-Vocabulary Object Detection upon Frozen Vision and Language Models},
  author={Kuo, Weicheng and Cui, Yin and Gu, Xiuye and Piergiovanni, AJ and Angelova, Anelia},
  booktitle={International Conference on Learning Representations},
  year={2023}
}

@inproceedings{kim2023region,
  title={Region-Aware Pretraining for Open-Vocabulary Object Detection with Vision Transformers},
  author={Kim, Dahun and Angelova, Anelia and Kuo, Weicheng},
  booktitle={Proceedings of the IEEE/CVF Conference on Computer Vision and Pattern Recognition},
  pages={11144--11154},
  year={2023}
}

@inproceedings{minderer2023scaling,
  title={Scaling Open-Vocabulary Object Detection},
  author={Minderer, Matthias and Gritsenko, Alexey and Houlsby, Neil},
  booktitle={Advances in Neural Information Processing Systems},
  volume={36},
  pages={72983--73007},
  year={2023}
}

@inproceedings{zhang2022dino,
  title={DINO: DETR with Improved DeNoising Anchor Boxes for End-to-End Object Detection},
  author={Zhang, Hao and Li, Feng and Liu, Shilong and Zhang, Lei and Su, Hang and Zhu, Jun and Ni, Lionel M. and Shum, Heung-Yeung},
  booktitle={International Conference on Learning Representations},
  year={2023}
}

@inproceedings{zhong2022regionclip,
  title={RegionCLIP: Region-Based Language-Image Pretraining},
  author={Zhong, Yiwu and Yang, Jianwei and Zhang, Pengchuan and Li, Chunyuan and Codella, Noel and Li, Liunian Harold and Zhou, Luowei and Dai, Xiyang and Yuan, Lu and Li, Yin and Gao, Jianfeng},
  booktitle={Proceedings of the IEEE/CVF Conference on Computer Vision and Pattern Recognition},
  pages={16793--16803},
  year={2022}
}

@inproceedings{luddecke2022clipseg,
  title={Image Segmentation Using Text and Image Prompts},
  author={L{\"u}ddecke, Timo and Ecker, Alexander S.},
  booktitle={Proceedings of the IEEE/CVF Conference on Computer Vision and Pattern Recognition},
  pages={7086--7096},
  year={2022}
}

@inproceedings{rao2022denseclip,
  title={DenseCLIP: Language-Guided Dense Prediction with Context-Aware Prompting},
  author={Rao, Yongming and Zhao, Wenliang and Chen, Guangyi and Tang, Yansong and Zhu, Zheng and Huang, Guan and Zhou, Jie and Lu, Jiwen},
  booktitle={Proceedings of the IEEE/CVF Conference on Computer Vision and Pattern Recognition},
  pages={18061--18070},
  year={2022}
}

@article{yu2021usv,
  title={{USV} Path Planning Method with Velocity Variation and Global Optimisation Based on {AIS} Service Platform},
  author={Yu, K. and Liang, X. F. and Li, M. Z. and Chen, Z. and Yao, Y. L. and Li, X. and Teng, Y.},
  journal={Ocean Engineering},
  volume={236},
  pages={109560},
  year={2021},
  publisher={Elsevier}
}

@article{singh2018constrained,
  title={A Constrained {A*} Approach towards Optimal Path Planning for an Unmanned Surface Vehicle in a Maritime Environment Containing Dynamic Obstacles and Ocean Currents},
  author={Singh, Y. and Sharma, S. and Sutton, R. and Hatton, D. and Khan, A.},
  journal={Ocean Engineering},
  volume={169},
  pages={187--201},
  year={2018},
  publisher={Elsevier}
}

@article{liu2025hybrid,
  title={Hybrid Path Planning Method for {USV} Based on Improved {A-Star} and {DWA}},
  author={Liu, Y. and Sun, Z. and Wan, J. and Li, H. and Yang, D. and Li, Y. and Sun, J.},
  journal={Journal of Marine Science and Engineering},
  volume={13},
  number={5},
  pages={934},
  year={2025},
  publisher={MDPI}
}

@article{schoener2022anytime,
  title={An Anytime Visibility--Voronoi Graph-Search Algorithm for Generating Robust and Feasible Unmanned Surface Vehicle Paths},
  author={Schoener, Michael and Coyle, Edward and Thompson, Derek},
  journal={Autonomous Robots},
  volume={46},
  number={8},
  pages={911--927},
  year={2022},
  publisher={Springer}
}

@article{wu2024efficient,
  title={Efficient Coverage Path Planning and Underwater Topographic Mapping of an {USV} Based on {A*}-Improved Bio-Inspired Neural Network},
  author={Wu, N. and Wang, R. and Qi, J. and Wang, Y. and Wen, G.},
  journal={IEEE Transactions on Intelligent Vehicles},
  year={2024},
  publisher={IEEE}
}

@article{luo2025lstm,
  title={Research on {LSTM-PPO} Obstacle Avoidance Algorithm and Training Environment for Unmanned Surface Vehicles},
  author={Luo, W. and Wang, X. and Han, F. and Zhou, Z. and Cai, J. and Zeng, L. and Zhou, X.},
  journal={Journal of Marine Science and Engineering},
  volume={13},
  number={3},
  pages={479},
  year={2025},
  publisher={MDPI}
}

@article{qu2025collaborative,
  title={The Collaborative Navigation Decision-Making Method of {USV} by {UAV} Based on Improved {PPO} Algorithm},
  author={Qu, S. and Guan, W. and Hu, T. and Cui, Z.},
  journal={Ocean Engineering},
  volume={341},
  pages={122381},
  year={2025},
  publisher={Elsevier}
}

@article{zhang2024multi,
  title={Multi-{USV} Task Planning Method Based on Improved Deep Reinforcement Learning},
  author={Zhang, J. and Ren, J. and Cui, Y. and Fu, D. and Cong, J.},
  journal={IEEE Internet of Things Journal},
  volume={11},
  number={10},
  pages={18549--18567},
  year={2024},
  publisher={IEEE}
}

@article{maidana2023risk,
  title={Risk-Based Path Planning for Preventing Collisions and Groundings of Maritime Autonomous Surface Ships},
  author={Maidana, R. G. and Kristensen, S. D. and Utne, I. B. and S{\o}rensen, A. J.},
  journal={Ocean Engineering},
  volume={290},
  pages={116417},
  year={2023},
  publisher={Elsevier}
}

@article{qiao2023survey,
  title={Survey of Deep Learning for Autonomous Surface Vehicles in Marine Environments},
  author={Qiao, Y. and Yin, J. and Wang, W. and Duarte, F. and Yang, J. and Ratti, C.},
  journal={IEEE Transactions on Intelligent Transportation Systems},
  volume={24},
  number={4},
  pages={3678--3701},
  year={2023},
  publisher={IEEE}
}

@inproceedings{khairuddin2015review,
  title={Review on Simultaneous Localization and Mapping ({SLAM})},
  author={Khairuddin, A. R. and Talib, M. S. and Haron, H.},
  booktitle={Proceedings of the 2015 IEEE International Conference on Control System, Computing and Engineering},
  pages={85--90},
  year={2015},
  organization={IEEE}
}

@article{lee2007constrained,
  title={A Constrained {SLAM} Approach to Robust and Accurate Localisation of Autonomous Ground Vehicles},
  author={Lee, K. W. and Wijesoma, S. and Guzm{\'a}n, J. I.},
  journal={Robotics and Autonomous Systems},
  volume={55},
  number={7},
  pages={527--540},
  year={2007},
  publisher={Elsevier},
  doi={10.1016/j.robot.2007.02.004}
}

@article{shalal2015orchard,
  title={Orchard Mapping and Mobile Robot Localisation Using On-Board Camera and Laser Scanner Data Fusion--Part {B}: Mapping and Localisation},
  author={Shalal, Nagham and Low, Tobias and McCarthy, Cheryl and Hancock, Nigel},
  journal={Computers and Electronics in Agriculture},
  volume={119},
  pages={267--278},
  year={2015},
  publisher={Elsevier},
  doi={10.1016/j.compag.2015.09.026}
}

@article{yang2024digital,
  title={Digital Twin-Based Autonomous Navigation and Control of Omnidirectional Mobile Robots},
  author={Yang, H. and Qin, Z. and Xia, Y. and Cheng, F.},
  journal={IEEE Transactions on Vehicular Technology},
  volume={74},
  number={4},
  pages={5687--5697},
  year={2024},
  publisher={IEEE}
}

@article{berg2025digital,
  title={Digital Twin Syncing for Autonomous Surface Vessels Using Reinforcement Learning and Nonlinear Model Predictive Control},
  author={Berg, H. S. and Menges, D. and Tengesdal, T. and Rasheed, A.},
  journal={Scientific Reports},
  volume={15},
  number={1},
  pages={9344},
  year={2025},
  publisher={Nature Publishing Group}
}

@article{xue2021semantic,
  title={Semantic Enrichment of Building and City Information Models: A Ten-Year Review},
  author={Xue, F. and Wu, L. and Lu, W.},
  journal={Advanced Engineering Informatics},
  volume={47},
  pages={101245},
  year={2021},
  publisher={Elsevier}
}

@techreport{iala2022g1132,
  author      = {{International Association of Marine Aids to Navigation and Lighthouse Authorities}},
  title       = {{IALA Guideline G1132: VTS Voice Communications and Phraseology}},
  institution = {International Association of Marine Aids to Navigation and Lighthouse Authorities},
  number      = {G1132},
  year        = {2022},
  note        = {Edition 2.2, revised on 31 January 2022}
}

@misc{imo2001smcp,
  author       = {{International Maritime Organization}},
  title        = {{IMO Standard Marine Communication Phrases}},
  howpublished = {IMO Resolution A.918(22)},
  year         = {2001},
  note         = {Adopted on 29 November 2001}
}

@article{xu2025llm4sac,
  author  = {Xu, Chenhang and Chu, Yijie and Gao, Qizhong and Wu, Ziniu and Wang, Jia and Yue, Yong and Dominik, Wojtczak and Zhu, Xiaohui},
  title   = {Autonomous Unmanned Surface Vehicle Docking Using Large Language Model Guide Reinforcement Learning},
  journal = {Ocean Engineering},
  volume  = {323},
  pages   = {120608},
  year    = {2025},
  doi     = {10.1016/j.oceaneng.2025.120608}
}

@article{christensen2025aicaptain,
  author  = {Christensen, Kim Alexander and Gusev, Alexey and Tufte, Andreas Gudahl and Alsos, Ole Andreas and Steinert, Martin},
  title   = {{AI Captain}: Conversational Mission Planning and Execution System for Autonomous Surface Vehicles},
  journal = {Ocean Engineering},
  volume  = {338},
  pages   = {121988},
  year    = {2025},
  doi     = {10.1016/j.oceaneng.2025.121988}
}

@article{salgado2026usv3,
  author  = {Salgado, Alex and Vasconcellos, Eduardo Charles and Guerra, Raphael and Gon{\c{c}}alves, Luiz Marcos Garcia and Clua, Esteban Walter Gonzalez},
  title   = {{USV-3.0}: Cognitive Maritime Navigation Through Vision-Language Models, Human-in-the-Loop Learning, and Spatio-Temporal Memory},
  journal = {Ocean Engineering},
  volume  = {355},
  number  = {1},
  pages   = {125010},
  year    = {2026},
  doi     = {10.1016/j.oceaneng.2026.125010}
}

@article{christensen2026foundation,
  author  = {Christensen, Kim Alexander and Tufte, Andreas Gudahl and Gusev, Alexey and Sinha, Rohan and Ganai, Milan and Alsos, Ole Andreas and Pavone, Marco and Steinert, Martin},
  title   = {Foundation Models on the Bridge: Semantic Hazard Detection and Safety Maneuvers for Maritime Autonomy with Vision-Language Models},
  journal = {Ocean Engineering},
  volume  = {359},
  number  = {3},
  pages   = {124646},
  year    = {2026},
  doi     = {10.1016/j.oceaneng.2026.124646}
}
\end{document}